\documentclass[journal,twoside,]{IEEEtran}
\ifCLASSINFOpdf
\else
\fi
\usepackage{url}

\usepackage{verbatim} 
\usepackage{graphicx}
\usepackage{setspace} 
\usepackage{tablefootnote}
\usepackage{color}
\usepackage[dvipsnames]{xcolor}

\begin{document}


%
\title{The CRAM Cognitive Architecture \\for Robot Manipulation in Everyday Activities}
%
%
%

\author{Michael~Beetz,~\IEEEmembership{Member,~IEEE,}
            Gayane Kazhoyan, 
           and~David~Vernon,~\IEEEmembership{Senior~Member,~IEEE}
\thanks{This work was supported by the German Research Foundation DFG, as part of the Collaborative Research Center (Sonderforschungsbereich) 1320 ``EASE --- Everyday Activity Science and Engineering'', University of Bremen (http://www.ease-crc.org). (\em{Corresponding author: David Vernon.})}
\thanks{The authors are with the Institute for Artificial Intelligence (IAI), University of Bremen, Am Fallturm 1, 28359 Bremen, Germany (e-mail: mbeetz@uni-bremen.de, gkazhoyan@uni-bremen.de, dvernon@uni-bremen.de).}
}

\maketitle

\begin{abstract}
This paper presents a hybrid robot cognitive architecture, CRAM, that enables robot agents to accomplish everyday  manipulation tasks. It addresses five key challenges that arise when carrying out everyday activities. These include (i) the underdetermined nature of task specification, (ii) the generation of context-specific behavior, (iii) the ability to make decisions based on knowledge, experience, and prediction, (iv) the ability to reason at the levels of motions  and sensor data, and  (v) the ability to explain actions and the consequences of these actions.  We explore the computational foundations of the CRAM cognitive model: the self-programmability entailed by physical symbol systems, the CRAM plan language, 
generalized action plans and implicit-to-explicit manipulation,  generative models, digital twin knowledge representation \& reasoning,  and narrative-enabled episodic memories. We describe the structure of the cognitive architecture and explain the process by which CRAM transforms generalized action plans 
into parameterized motion plans. It does this using  knowledge and reasoning to identify the parameter values that maximize the likelihood of successfully accomplishing the  action. We demonstrate the ability of a CRAM-controlled robot to carry out everyday activities in a kitchen environment.
Finally, we consider future extensions that focus on achieving greater flexibility through transformational learning and metacognition.
\end{abstract}


%
\IEEEpeerreviewmaketitle

\section{Cognitive Architectures}
\label{section:cognition}
%
%
%
%
\IEEEPARstart{T}{he} concept of a cognitive architecture, introduced by Allen Newell \cite{Newell90}, arose from over sixty years of research in various strands of cognitive science, a discipline that embraces  neuroscience, cognitive psychology, linguistics, epistemology, philosophy, and artificial intelligence, among others. The primary goal of cognitive science is to explain the underlying processes of human cognition, ideally in the form of a  model that can be replicated in artificial agents. It has its roots in cybernetics \cite{Wiener48}, but appears as a formal discipline referred to as cognitivism in the late 1950s. Cognitivism takes a computational stance on cognitive function  and uses symbolic information processing as its core model of cognition and intelligence \cite{NewellSimon76}.  Cybernetics also gave rise to the alternative  emergent systems approach which recognized the importance of self-organization in cognitive processes, eventually embracing connectionism, dynamical systems theory, and enaction \cite{StewartGapenneDiPaolo10}. Hybrid systems seek to combine  the cognitivist and emergent approaches to varying degrees in an effort to exploit the knowledge representation and reasoning of symbolic approaches with sub-symbolic representation and inference.
 
A cognitive architecture is a software framework that integrates all the elements required for a system to exhibit the abilities that are considered to be characteristic of a cognitive agent.  Core cognitive abilities include perception, action, learning, adaptation, anticipation \& prospection, motivation, autonomy, internal simulation, attention, action selection, memory, reasoning, and meta-reasoning \cite{KotserubaTsotsos20,VernonvonHofstenFadiga16}. A cognitive architecture  determines the overall structure and organization of a cognitive system, including the component parts or modules \cite{Sun04}, the relations between these modules, and the essential algorithmic and representational details within them \cite{Langley06}. 

There are three different types of cognitive architecture, each derived from the three approaches to cognitive science: the cognitivist, the emergent, and the hybrid. 

Cognitivist cognitive architectures, often referred to as symbolic cognitive architectures \cite{KotserubaTsotsos20}, focus on the aspects of cognition that are relatively constant over time and that are independent of the task \cite{RitterYoung01,LangleyLairdRogers09}, with knowledge providing the task-specific element. The combination of a cognitive architecture and a particular knowledge set is referred to as a cognitive model.  In many cognitivist systems, much of the knowledge incorporated in the model is  provided by the designer, possibly drawing on years of experience working in the problem domain. Machine learning is increasingly used to augment and adapt this knowledge.

Emergent cognitive architectures focus on the development of the agent from a primitive state to a fully cognitive state over its life-time. As such, an emergent cognitive architecture is both the initial state from which an agent subsequently develops and the encapsulation of the  dynamics that drive that development,  typically exploiting sub-symbolic  processes and representations.  Since the emergent paradigm holds that the body of the cognitive agent plays a causal role in the cognitive process, emergent cognitive architectures try to reflect in some way the structure and capabilities of the physical body and its morphological development \cite{Naya-VarelaFainaDuro2021}.

Hybrid systems endeavour to combine the strengths of the cognitivist and emergent approaches. Most hybrid systems focus on integrating symbolic and sub-symbolic processing.  Hybrid cognitive architectures are the most prevalent type: forty-eight of the eighty-four cognitive architectures surveyed by Kotseruba and Tsotsos \cite{KotserubaTsotsos20} are hybrid. 

Most cognitive architectures focus on modelling  human cognition and many are  a candidate unified theory of cognition \cite{Newell90}, e.g 
Soar \cite{LairdNewellRosenbloom87,Laird12}, 
ACT-R \cite{Anderson96,Andersonetal04}, 
CLARION \cite{Sun07,Sun16,Sun17}, and 
LIDA \cite{RamamurthyBaarsDMelloFranklin06,Franklinetal14},   all of which are classified as hybrid by Kotseruba and Tsotsos \cite{KotserubaTsotsos20}.
While some of these cognitive architectures have been applied in robotics, e.g. Soar and LIDA, and ACT-R/E \cite{Traftonetal07}, other cognitive architectures have been designed specifically in the context of robotics research, e.g. ArmarX \cite{Vahrenkampetal2015},  CRAM \cite{BeetzMosenlechnerTenorth10}, and ISAC \cite{Kawamuraetal08}, and do not claim to be unified theories of cognition.  Nevertheless, robot cognitive architectures and cognitive architectures that model human cognition  both share a focus on achieving the versatility that humans exhibit, for example when engaged in everyday activities, the subject of the next section.

\begin{figure}[tb]
  \centering
  \includegraphics[width=\columnwidth]%
  {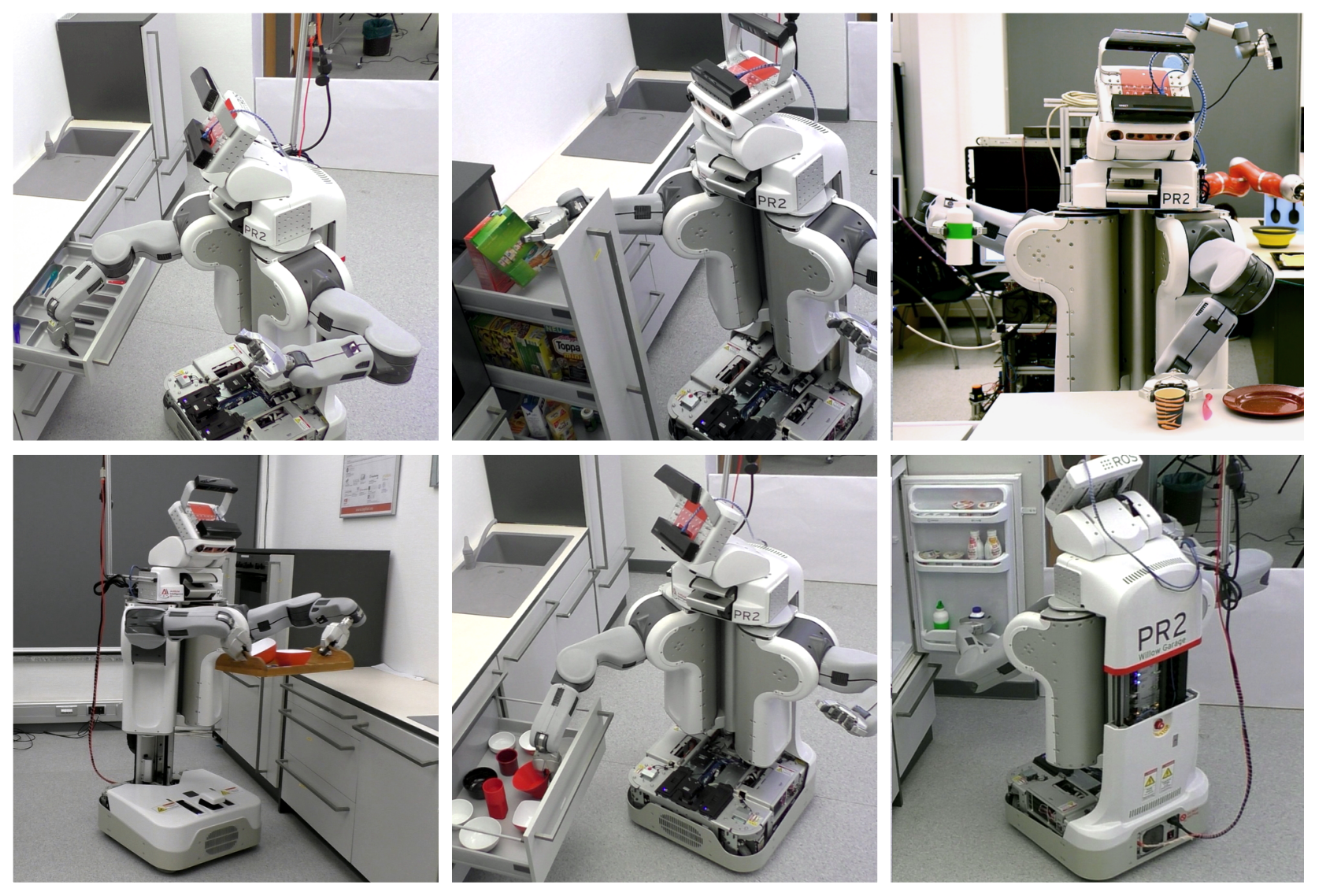}
  \caption{Different object grasps selected by the generative model based on the object, task, and context.}
  \label{fig:blackbox-mastery}
\end{figure}
\vspace{-2mm}

\section{Challenges of Manipulation Tasks in Everyday Activities}
\label{section:challenges}

The previous section identified action selection and action execution as core abilities of cognitive architectures. In the context of cognitive robotics, this translates to the need for robot agents to be able to accomplish  everyday manipulation tasks and explain how they accomplish them. Consider the task of setting a table for a meal and tidying up afterwards, shown in Figure~\ref{fig:blackbox-mastery} and a video recording of  this an activity.\footnote{\label{footnote:video} \scriptsize\url{https://www.ease-crc.org/link/video-ease-robot-day.}}  The robot fetches the required items and arranges them in the expected layout on the table. To accomplish the task successfully the  robot has to select an appropriate behavior for every object  transportation task, depending on the type and state of the object to  be transported (be it a spoon, bowl, cereal box, milk box, or mug), the original location (be it the drawer, the high drawer, or the table), and the task context (be it setting or cleaning the table, loading the dishwasher, or throwing away items). As the behavior is not constrained by the task request, the robot has to infer the appropriate behavior using its knowledge and reasoning capabilities.  This reveals that an everyday activity is typically a common complex  task about which an agent  has considerable knowledge, often making  the task mundane, especially if it only has to be accomplished in a satisfactory manner, rather than in an optimal manner \cite{anderson95phd}. Nevertheless, carrying out everyday activities poses five key challenges for cognitive robotics.

First, requests for accomplishing everyday tasks are typically underdetermined.  Requests such as ``set the table,'' ``load the   dishwasher,'' and ``prepare breakfast'' typically do not fully specify  the intended goal state, yet the requesting agent has specific  expectations about the results of the activity.  Consequently, the robot agent carrying  out the everyday activity needs to acquire  the missing knowledge to accomplish the task and meet those expectations.  

Second, accomplishing everyday tasks requires context-specific  behavior. The underdetermined request ``fetch some object and put it where it belongs'' has to generate   different behaviors depending on the object, its state, the current
location of the object, the scene context, the destination of the  object, and the task context. The behavior has to be carefully
chosen to match the current contextual conditions. Variations in these conditions require adaptive behavior.
  
Third, competence in accomplishing everyday manipulation tasks  requires the ability to make decisions based on knowledge, past experience, and prediction of the outcome of each constituent  action. The knowledge required includes common sense, such as  knowing that the tableware to be placed on the table should be clean and that clean tableware is typically stored in cupboards. It also requires intuitive physics knowledge, e.g. that objects should be placed with their center of gravity close to the support surface to avoid them toppling over.  Domain knowledge might include the fact that plates are made of porcelain, which is a breakable material, so they must be manipulated with care.  Experience allows the robot agent to improve the robustness and efficiency of its actions by tailoring behavior to specific contexts. Prediction enables the robot to take likely consequences of actions into account, such as predicting that using a specific grasp would require the object to be subsequently re-grasped in order to place it at the intended location. 

Fourth, accomplishing everyday manipulation tasks requires the robot agent to reason at the motion level, predicting the way the parameterization of motions alters the physical effects of these motions,  and thereby identifying the best way to achieve the intended outcomes and avoid unwanted side effects.  For example, in order to pour something from a pot onto a plate, a robot agent has to infer that it has initially to hold the pot  horizontally and then tilt it. To do this, it may have to grasp the pot with two hands, such that the center of mass is in between the hands, because the tilting motion is the easiest when rotating the pot around the axis between the handles. 

Fifth, a robot agent should be able to answer  questions about what it is doing, why it is doing it, how it is doing it, what it expects to happen when it does it, how it could do it differently, what are the advantages and disadvantages of doing it one way or another, and so on. 
This ability is important because it allows the robot agent to assess the impact of not doing something effectively or failing to do it at all. This understanding also allows a robot agent  to discover possibilities for improving the way it currently does things.  

In the following, we explain how the CRAM cognitive architecture addresses these five challenges, first by stating the guiding principle of the CRAM generative model of robot agency and discussing the four core elements of this model, second by describing the structure and operation of the CRAM cognitive architecture, and third by providing examples of a CRAM-controlled robot carrying out manipulation tasks in everyday activities in a kitchen.

\section{The CRAM Generative Model of Robot Agency}
\label{sec:ease-approach}

\subsection{Background and Computational Foundations}
\label{section:foundations}

\begin{figure}[t]
\unitlength 1mm
\begin{center}
\includegraphics[height=45mm,angle=0]{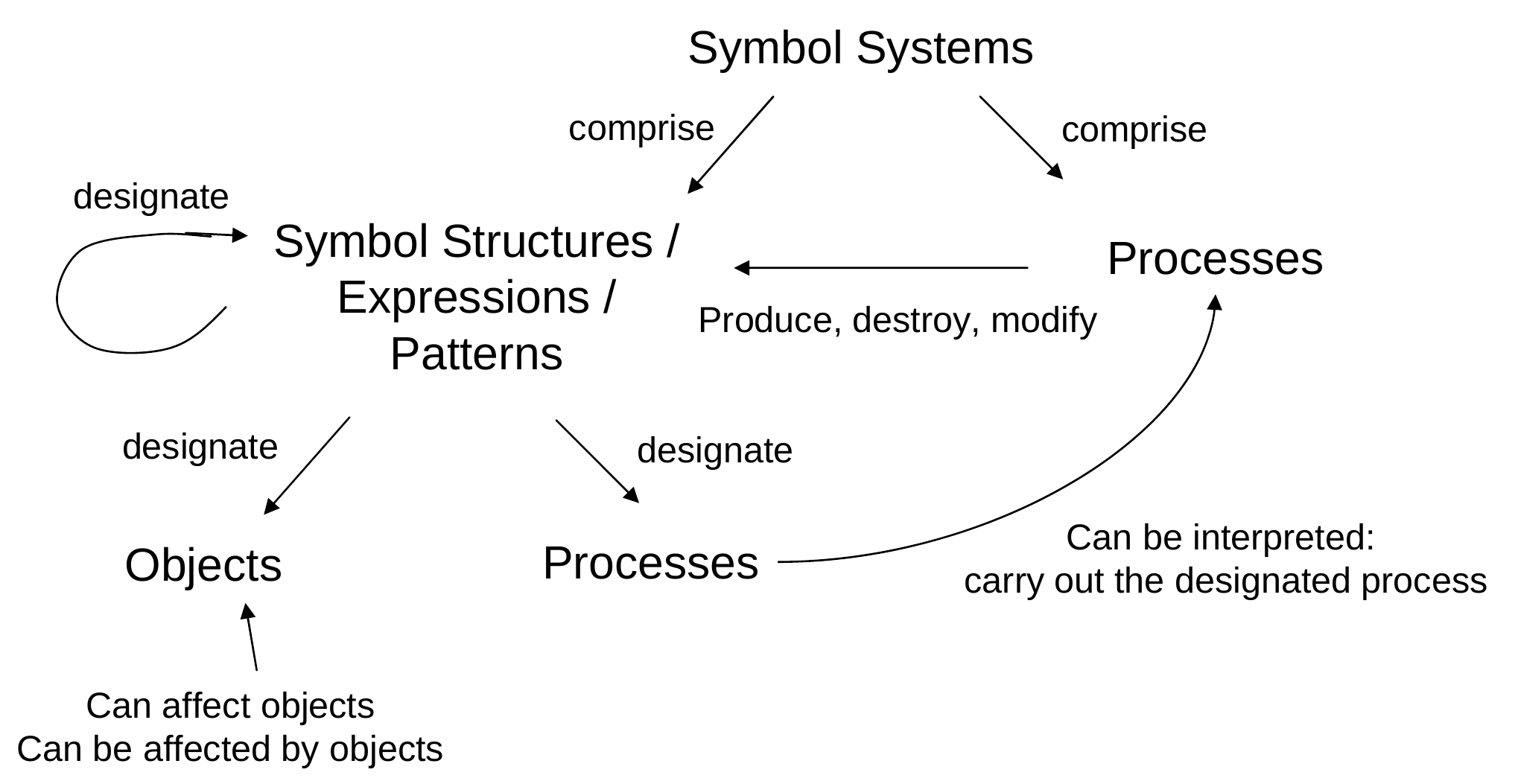}
\end{center}
\caption{The essence of a physical symbol system (from \protect\cite{VernonMettaSandini07}).}
\label{fig:physical_symbol_system}       
\end{figure}

For the design of CRAM, we take the stand point that physical symbol systems as proposed by Newell and Simon \cite{NewellSimon76} can be harnessed as essential computational tools for
intelligent agency.
Symbols are a physical patterns and a physical symbol system (see Fig. \ref{fig:physical_symbol_system}) is a machine that produces over time an evolving collection of symbol structures, i.e.  expressions.  The system also comprises processes that operate on expressions to produce other expressions: to create, modify, reproduce, and destroy them.  An expression can designate an object and thereby the system can either affect the object itself or behave in ways that depend on the object. An expression can also designate a process, in which case   the system can interpret the expression by carrying out the process.   

This definition of a symbol system by Newell and Simon \cite{NewellSimon76}  involves a crucial element of circularity: processes can produce patterns, and patterns can designate  processes (which can be interpreted).  Thus, not only can a physical symbol system construct representations of objects and reason about those representations, but it can also reason about the processes that produce and interpret those representations, and it can modify those processes.  In other words, physical symbol systems have an intrinsic capacity for autonomous development through self-programming, at least in principle. 

CRAM is founded on this  self-programming capacity of physical symbol systems, adopting the perspective of {\em program as data object}, and, specifically, {\em program as inspectable and transformable data object}. It takes the stance that robot agents are controlled by programs (i.e. processes designated by expressions),  formally represented symbol structures that prescribe the computations for inferring the body motions to be executed by the robots. CRAM represents parts of these programs as plans, which we take to   be parts of the control program that can be both executed (i.e. interpreted) and reasoned about and manipulated at execution time.\footnote{Parts of the control program can also be implemented  as distributed representations such as learned artificial neural networks (ANNs) and policies.  CRAM considers ANNs and policies as blackbox approximations of functions given by a set of training  examples that optimize a given objective function.}

In this section, we describe the plan language in which key components of the program are expressed, in particular generalized action plans and motion plans, how motions are computed from underdetermined  action descriptions, and the concept of generative models that capture how preference relations over motions are implemented, i.e. how specific and appropriate motions are selected when interpreting the generalized action plan.  We defer a discussion of the  two remaining elements of the computational foundations of the CRAM cognitive model ---  digitial twin knowledge representation \& reasoning (DTKR\&R) and narrative-enabled episodic memories (NEEMs) --- to Section \ref{sec:CRAM_CA} where we describe the CRAM cognitive architecture that realizes this model and the role that DTKR\&R and NEEMs play in that realization.

\subsection{Overview}

The guiding principle of CRAM is the need for cognitive robots to achieve {\em implicit-to-explicit} manipulation. The term implicit-to-explicit is used to convey the idea that the  manipulation task is accomplished by mapping a vaguely-stated  high-level goal  to the specific low-level motions required to accomplish the goal. It also includes the idea that any constraints that arise at the high  level are propagated to the low-level execution in that mapping.   Implicit-to-explicit also means that  success in accomplishing  the task is evaluated in the same space as that in which the goal was formulated, i.e.\ the perceptual space comprising observations of the  world before starting the task and observations of the world  after completing it, and every instant in between.

This guiding principle motivated the creation of an activity description programming language --- the CRAM Plan Language (CPL) --- which allows manipulation actions to be carried out successfully  by saying {\em what} action is to be carried out, but without having to say {\em how}  it has to be carried out. We refer to this as an {\em  underdetermined action description}: a high-level abstract specification of the robot actions required to carry out an everyday task. This action specification is framed in incomplete terms, i.e. it doesn't  provide all knowledge required to complete the task.   Such vaguely-stated instructions are what people typically give when they  ask someone to do something, for example ``fetch the milk and pour it in the glass.''  The knowledge required to complete the action is acquired when the underdetermined action description is being performed.  This approach uses  reasoning and contextual knowledge  to identify the information that is missing in the underdetermined action description, and inference and prospection to identify and validate the robot motions that are most likely to result in the successful execution of the required action.   In CRAM, as we will see below, prospection is achieved using a high-fidelity virtual reality  physics engine simulation of a digital twin of the robot and its environment.

In CPL, actions are decomposed into primitive motions, each of which has a set of parameters, the values of which determine the exact nature of the motion.  CRAM uses a {\em generative model} to map from desired outcomes of an action to the motion parameter values that are most likely to succeed in accomplishing the desired action.  CPL provides for the creation of a {\em generalized action plan}:  a high-level plan which is expanded into a low-level parameterized motion plan, supplied with appropriate parameter values, and then executed.  As a computational expression of the underdetermined action description, a generalized action plan is also underdetermined: not all the knowledge required to execute the plan is specified.  The required knowledge takes the form of the values of the parameters of the motion primitives into which the generalized action plan is expanded.  As we have said, these motion parameter values maximize the likelihood that  the associated body motions  successfully accomplish the desired action.  These values  are provided by the generative model.  Expansion of the generalized action plan and identification of motion parameter values is  a process referred to as {\em contextualization}. It operates on an element of the generalized action plan called an {\em action designator}, a placeholder for yet-to-be-determined information. The designator is resolved and the related information is determined at run time  based on the current context of the task action.    

\begin{figure}[tb]
  \centering
  \includegraphics[width=1.0\columnwidth]
  {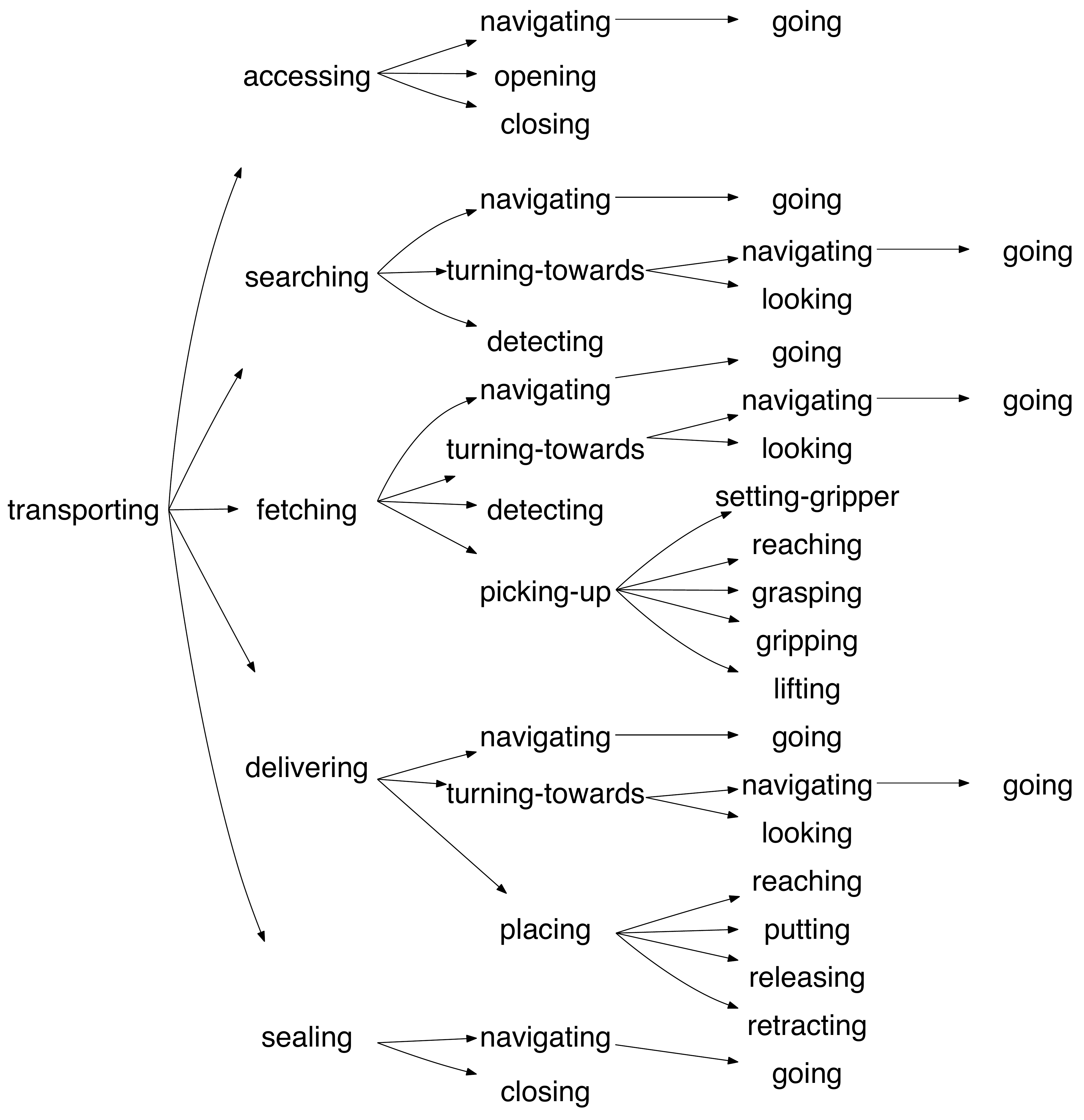}  
  \caption{The hierarchy of action designators for the PR2 robot.
Note that the resolution of an action hierarchy at one level may cause  the instantiation of more than one instance of a given action designator at the next level. Here, for the sake of compact presentation,  we show just a single instance.  Action designators that are leaf nodes in this tree are referred to as atomic action designators. These are subsequently resolved into  motion designators; see Table \ref{table:motion-designators}.}
  \label{fig:action_designator_hierarchy}
\end{figure}

There are four types of designator in CPL: action, object, location, and motion designators.    Action designators  and motion designators are both concerned with robot control. Action designators focus on achieving some goal state, e.g. having set the table for a meal or having placed dirty dishes in the dishwasher, whereas motion designators are concerned with the physical movements and the control of some actuator, e.g.  setting the gaze direction of the robot head, moving the end-effector to a given pose, or actuating the gripper.  Location designators are concerned with poses in general, e.g. a list of positions and orientations where a robot might direct its gaze when looking for some specific object or a list of positions and orientations at which the robot should stand in order to perform some manipulation task. Object designators are concerned with the properties of objects in the robot's environment, e.g. the pose of the object and its physical characteristics. Object designators provide an interface to the perception system since the pose of an object will typically be determined at run time.

\renewcommand{\arraystretch}{0.85}
 
\begin{table}[t]
    \caption{Atomic action designators and their corresponding\protect\\motion designators for the PR2 robot.
}
    \label{table:motion-designators}
    \centering
    \begin{tabular}{|c|c|}
    \hline
       \textbf{\footnotesize \textsf{Action Designator}} & \textbf{\footnotesize \textsf{Motion Designator}}\\
    \hline
        {\footnotesize \textsf{closing}	}& {\footnotesize \textsf{closing }} \\
        {\footnotesize \textsf{detecting}} & {\footnotesize \textsf{detecting}} \\
        {\footnotesize \textsf{going}} & {\footnotesize \textsf{going}} \\
        {\footnotesize \textsf{grasping}} & {\footnotesize \textsf{moving-tcp}} \\
        {\footnotesize \textsf{gripping}} & {\footnotesize \textsf{gripping}} \\
        {\footnotesize \textsf{lifting}}	& {\footnotesize \textsf{moving-tcp}}\\
        {\footnotesize \textsf{looking}}	& {\footnotesize \textsf{looking}}\\
        {\footnotesize \textsf{--}} & {\footnotesize \textsf{moving-torso}}\\
        {\footnotesize \textsf{opening}}	& {\footnotesize \textsf{opening}}\\
        {\footnotesize \textsf{--}} & {\footnotesize \textsf{moving-arm-joints}}\\
        {\footnotesize \textsf{pulling}} & {\footnotesize \textsf{moving-tcp}}\\
        {\footnotesize \textsf{pushing}} & {\footnotesize \textsf{moving-tcp}}\\
        {\footnotesize \textsf{putting}} & {\footnotesize \textsf{moving-tcp}}\\
        {\footnotesize \textsf{reaching} }& {\footnotesize \textsf{moving-tcp}}\\
        {\footnotesize \textsf{releasing} }& {\footnotesize \textsf{opening}}\\
        {\footnotesize \textsf{retracting}} & {\footnotesize \textsf{moving-tcp}}\\
        {\footnotesize \textsf{setting-gripper}} & {\footnotesize \textsf{moving-gripper-joint}}\\
    \hline
    \end{tabular}
\vspace{-2 mm}
\end{table}
 
\renewcommand{\arraystretch}{1}

There is a hierarchy of action designators; see Fig. \ref{fig:action_designator_hierarchy}. Consequently, the resolution of an action designator can involve the instantiation and resolution of other action designators. The action designators at the lowest level in the hierarchy are referred to as atomic action designators. Ultimately, all action designators are resolved into more concrete motion, location, and object designators.   Specifically, atomic action designators are resolved directly into the motion designators that  form the motion plan; see Table \ref{table:motion-designators}. Designator resolution is accomplished either by querying  knowledge embedded in the plan,  by querying knowledge in CRAM's knowledge base, or by accessing sensorimotor data through perception.  Resolving a motion designator results in motion of the robot body.  

To summarize: there are three complementary perspectives on successfully accomplishing an action with CRAM. We speak of performing an {\em underdetermined action description},  interpreting a {\em generalized action plan}, and resolving the {\em high-level action designator} in a generalized action plan. The three terms are effectively equivalent. We use them in different
contexts: performing an underdetermined action description when focussing on the action, interpreting a generalized action plan when focussing on the way action is accomplished with the aid of the generative model, and resolving a high-level action designator when focussing on the expansion of the action designator into its constituent low-level designator components. 
Ultimately,  the process of refining an underdetermined action description, i.e.\ contextualizing a generalized action plan, is equivalent to the process of  resolving the high-level action designator. Associated with these three perspectives on a high-level action, there are three other elements: (i)  the corresponding low-level parameterized motion plan, (ii)   the contextualization process of mapping from the generalized action plan to this motion plan,  (iii) the generative model used to identify the motion plan parameter values that maximize the likelihood of a successful action.

Let us now look more closely at these four key elements: the high-level generalized action plan, the low-level motion plan,  the process of contextualization, and the generative model.

\subsection{Generalized Action Plans}
\label{section:generalized-action-plans}

A CRAM-based  robot agent is equipped with a  generalized action plan for each action category, which typically corresponds to action verbs such as \textsf{\footnotesize fetch}, \textsf{\footnotesize  place}, \textsf{\footnotesize pour}, and \textsf{\footnotesize cut}. A generalized action plan specifies an action schema, i.e.\  a template of how actions of this category can be executed. A generalized action plan is invoked with a request to perform an underdetermined action description and, in turn, a generalized action plan is interpreted by  resolving a high-level action designator in the generalized plan.  
A generalized action plan has the following generic form.
\vspace{-4mm}
\singlespace
\textsf{\footnotesize
\begin{tabbing}
~~~~~~~~\=~~~\=~~~\=~~~\=~~~~~~~~~~~~~~~~~\=~~~\=~~~\=~~~\=\\
\+\>                    ({\bf \textsf{def-plan}} some-generalized-action-plan ($\langle$list of parameters$\rangle$) \\
\> \>                      ({\bf \textsf{when}} ($\langle$some-required-preconditions-are-satisfied$\rangle$) \\
\> \> \>                    ({\bf \textsf{perform}} ({\bf \textsf{an action}} \\
\> \> \> \>                                  (type ?action-category) \\
\> \> \>  \>                                 ($\langle$key-1$\rangle$ $\langle$value-1$\rangle$) \\
\> \> \>  \>                                 ($\langle$key-2$\rangle$ $\langle$value-2$\rangle$) \\
\> \> \>  \>                                 \ldots  ))))\\
\end{tabbing}
}
\vspace{-2mm}
\noindent Some of the values of the key-value pairs will be provided as input by the generalized action plan parameter values, others will take the form of a constituent designator, e.g. a lower-level action designator, a motion designator, a location  designator, or an object designator.  Consider the following outline of the \textsf{\footnotesize fetch\&place} generalized action plan.
\vspace{-4mm}
\singlespace
\textsf{\footnotesize
\begin{tabbing}
~~~~~~~~\=~~~\=~~~\=~~~\=~~~~~~~~~~~~~~~~~\=~~~\=~~~\=~~~\=\\
\+\>\                 ({\bf \textsf{def-plan}} fetch\&place (\=?object-to-be-fetched, ?destination)\\
\> \>                      ({\bf \textsf{with-robot-at-location}} (?location-at-which-to-fetch) \\
\> \> \>                    ({\bf \textsf{perform}} ({\bf \textsf{an action}} \\
\> \> \> \>                                   (type picking-up) \\
\> \> \>  \>                                  (object ({\bf \textsf{an object}} (type ?object-to-be-fetched)))\\
\> \> \>  \>                                  (arm ?arm-to-be-used) \\
\> \> \>  \>                                  (grasp ?grasp-pose) \\
\> \> \>  \>                                  (lift-pose ?lift-pose-to-be-used)))\\
~\\
\> \>                      ({\bf \textsf{with-robot-at-location}} (?location-at-which-to-place) \\
 \> \> \>                    ({\bf \textsf{perform}} ({\bf \textsf{an action}} \\
\> \> \> \>                                   (type placing) \\
\> \> \>  \>                                  (object ({\bf \textsf{an object}} (type ?object-to-be-fetched)))\\
\> \> \>  \>                                  (target ({\bf \textsf{a location}} (pose ?destination))) \\
\> \> \>  \>                                  (arm ?arm-to-be-used) \\
\> \> \>  \>                                  (grasp ?grasp-pose) \\
\> \> \>  \>                                  (lower-pose ?lower-pose-to-be-used)))))\\
\end{tabbing}
}
\vspace{-3mm}
\noindent The generalized action plan  above is an action schema which provides a template for how to transport any object from  its current location, which might not be known, to any destination.  The plan language lets the programmer state underdetermined action descriptions which have to be contextualized by the robot agent at execution time. For example, in the underdetermined description of the \textsf{\footnotesize fetching} action, it is not specified with which arm to grasp the object, with what kind of a grasp, or with which force to squeeze the object when grasping it. It is also not stated where the object is located in the environment. All these  parameters have to be inferred from the robot's knowledge and supplemented at run time with perceptual information.

The {\bf\textsf{\footnotesize with-robot-at-location}} construct of the plan language ensures that  the robot is located appropriately during the action. What is appropriate depends on the object, the robot capabilities, the surroundings, and the task context. For example, when performing the \textsf{\footnotesize fetching} action, the robot should stand in a location where it is able to perceive and reach the object. If the robot is not at the appropriate location, performing the action is suspended, the robot repositions itself, and only then continues the action.

The body of the plan specifies the high-level logical structure of the plan. The plan body consists of two steps. The first step tells the robot to fetch a specific object (while standing at an appropriate location for fetching). The second step tells the robot that the object has to be placed at the given destination, and that the robot should ensure that it is standing at the appropriate location from where the destination is reachable.  Both steps comprise a high-level action designator, the first of type \textsf{\footnotesize picking-up}, the second of type \textsf{\footnotesize placing}.  When the generalized action plan is interpreted, these high-level action designators are resolved and the associated parameters are inferred by the CRAM cognitive architecture using a generative model. We note that in both action designators, one of the parameters is an object designator. Resolving this designator for the object to be fetched (passed as an argument to the generalized action plan) yields information about the object, e.g. its pose and physical properties.  Note too that, in the second \textsf{\footnotesize placing} action designator, the value part of the \textsf{\footnotesize target} key  is a location designator. Resolving this designator yields a list of candidate poses for \textsf{\footnotesize placing} the object.  

The underdetermined nature of the plan means that if the location of the object is not known then the robot has to search for it. The search process will be more efficient if the robot knows places where the object is likely to be.  The robot can infer this knowledge from its knowledge base. Similarly, the manner in which the object has  to be picked up can only be decided after the object is found and its geometry and state as well as the scene context are known.

\subsection{Motion Plans}
\begin{figure}[tb]
 \centering
 \includegraphics[width=\linewidth]{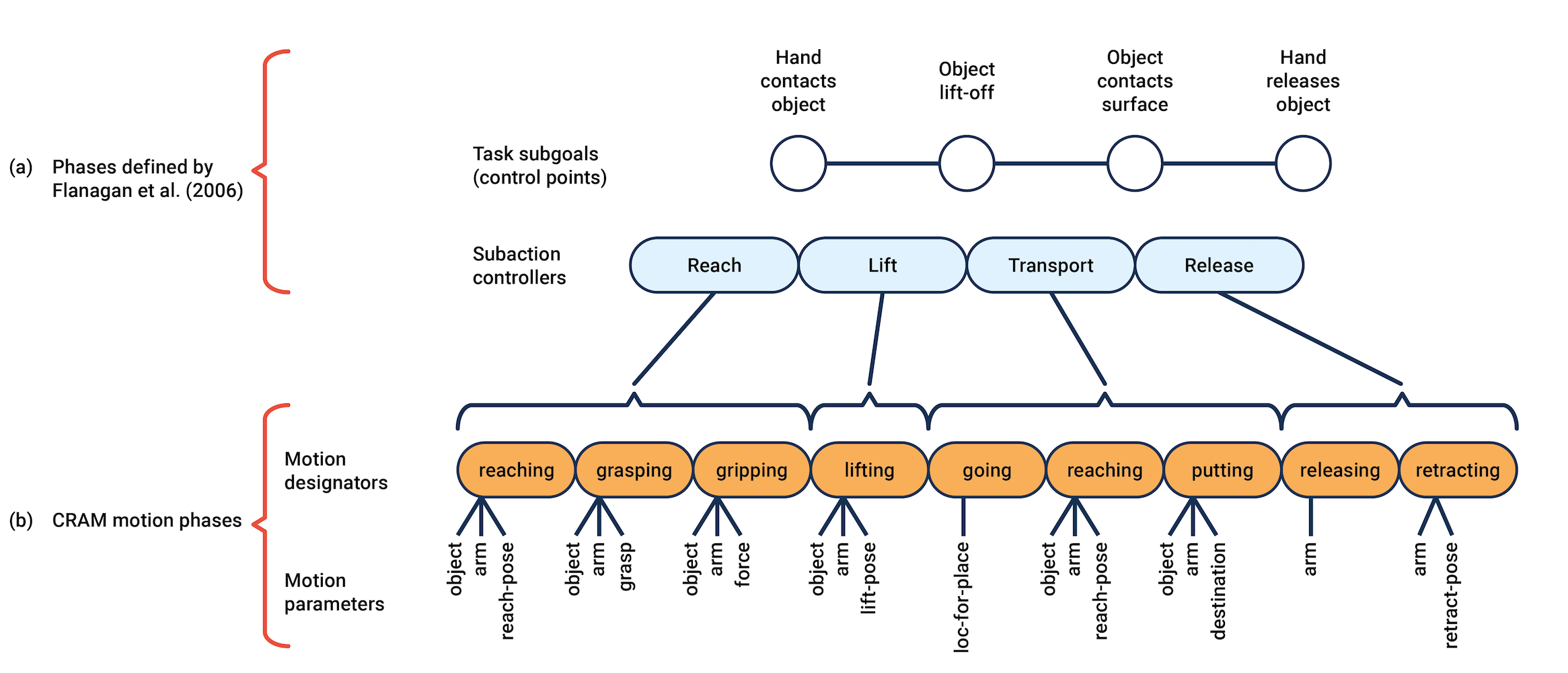}
\caption[]{Motion plan for  the \textsf{\footnotesize fetch\&place} generalized action plan. 
(a)  The motion phases of a pick and place task as described by Flanagan et al.
\cite{Flanaganetal2006}: 
motion phases, each governed by a  subaction controller,  are depicted in blue boxes while the control points depicted by circles represent the task subgoals. 
(b) The realization of this approach in CRAM, showing the  motion designators that comprise  each motion phase depicted in orange boxes, together with their associated motion parameters.}
\label{fig:generalized-motion-plan}
\end{figure}

The resolution of high-level action designators generates  motion schemata that tell the plan executive how to execute the constituent motions.  The motion schemata are represented as motion plans that structure the motion into motion phases. We have adopted an approach similar to that described by Flanagan et al. \cite{Flanaganetal2006}, but with a slightly finer-grained representation of motion primitives. For example, the motion plan for picking up an already detected object within the robot's reach and placing it at another location is shown in Fig.~\ref{fig:generalized-motion-plan}.

A  motion plan for fetch \& place actions comprises four different phases: reaching, lifting, transporting, and releasing \cite{Flanaganetal2006}. While the paper by Flanagan et al. \cite{Flanaganetal2006} refers to {\em action phase}, we use {\em motion phase} here to connote a lower level of abstraction and avoid confusion with the higher level of abstraction entailed by a generalized action plan and an action designator. Each motion phase has a goal.  When the goal is achieved, the start of the subsequent motion phase is triggered. Goals can be force-dynamic events, e.g.  the robot finger coming into contact with the object to be grasped, or a perceptually distinctive event, e.g. a milk carton becoming visible when a fridge door is opened. Each motion phase comes with several parameters that can be set in order to match the motion to the current situation. 

We note again that, with the CRAM model, it is assumed that for each action category we can specify a motion plan schema with a small number of motion parameters, that this motion plan is sufficient to achieve the desired outcome and avoid unwanted side effects, and that this plan can be used for a large variety of objects and tasks by leveraging the constraints imposed by the current context.

\subsection{Contextualization}
The action associated with an underdetermined action description is performed by first selecting  the generalized action plan for the action category corresponding to the action description.  Mapping from a generalized action plan to a motion plan is accomplished by the contextualization process. It  has three steps, as follows.

\begin{enumerate}
\item Instantiate the selected generalized action plan by inserting  the arguments required for the specific action to be performed.  One of these arguments is the action type, e.g. \textsf{\footnotesize fetch}, \textsf{\footnotesize place}, \textsf{\footnotesize fetch\&place},  \textsf\footnotesize {pick up}, \textsf{\footnotesize pour}, or \textsf{\footnotesize cut}. Others  include  the type of the object to be manipulated or the destination location.  These arguments are typically  designators of some kind, e.g. an action, object, or location designator.  
\item Extend the instantiated generalized action plan by adding the parameters needed to execute the motion plan, e.g. the arm or grasp pose to use. These motion parameters are identified by resolving the high-level action designator 
into the motion plan's constituent motion designators, by way of the action designator hierarchy; see Fig. \ref{fig:action_designator_hierarchy}.
\item Generate and submit  a query for the values of the motion parameters that will produce the robot movements required to achieve the goal.  
\end{enumerate}
The query that results from the third step is answered by a knowledge representation \& reasoning (KR\&R) system. The KR\&R system plays the role of the generative model,  exploiting the constraints of contextual knowledge and current perceptual 
information, and prospection, using the robot's inner world to  simulate plan execution. The answer to the query identifies the motion parameter values that produce  robot body motions that are most likely to succeed in accomplishing  the desired action.  For instance, a query for the \textsf{\footnotesize fetch\&place} example above might be formulated as follows.
\vspace{-5 mm}
\singlespace
\textsf{\footnotesize
\begin{tabbing}
~~~~~~~~\=~~~\=~~~\=~~~\=~~~~~~~~~~~~~~~~~\=~~~\=~~~\=~~~\=\\
\+\>\=({\bf \textsf{qu}}\={\bf\textsf{ery-variables}} (\=?location-at-which-to-fetch, \\
\> \> \>                                                 ?arm-to-be-used,    ?grasp-pose, \\
\> \> \>                                                 ?lift-pose-to-be-used, ?location-at-which-to-place, \\
\> \> \>                                                 ?lower-pose-to-be-used)\\
\> \>                                 {\bf \textsf{to-succeed}} (\\
~~~~~~~~\=~~~\=~~~\=~~~\=~~~~~~~~~~~~~~~~~\=~~~\=~~~\=~~~\=\\
\> \>                      ({\bf \textsf{\footnotesize with-robot-at-location}} (?location-at-which-to-fetch) \\
\> \> \>                    ({\bf \textsf{perform}} ({\bf \textsf{an action}} \\
\> \> \> \>                                   (type picking-up) \\
\> \> \>  \>                                  (object ({\bf \textsf{an object}} (type ?object-to-be-fetched)))\\
\> \> \>  \>                                  (arm ?arm-to-be-used) \\
\> \> \>  \>                                  (grasp ?grasp-pose) \\
\> \> \>  \>                                  (lift-pose ?lift-pose-to-be-used)))\\
~\\
\> \>                      ({\bf \textsf{with-robot-at-location}} (?location-at-which-to-place) \\
 \> \> \>                    ({\bf \textsf{perform}} ({\bf \textsf{an action}} \\
\> \> \> \>                                   (type placing) \\
\> \> \>  \>                                  (object ({\bf \textsf{an object}} (type ?object-to-be-fetched)))\\
\> \> \>  \>                                  (target ({\bf \textsf{a location}} (pose ?destination))) \\
\> \> \>  \>                                  (arm ?arm-to-be-used) \\
\> \> \>  \>                                  (grasp ?grasp-pose) \\
\> \> \>  \>                                  (lower-pose ?lower-pose-to-be-used))))\\
\end{tabbing}
} 
\noindent Once the query has been answered  by the KR\&R  system and the parameter values have been  determined, the motion plan is executed adaptively by the action executive.

Choosing the appropriate motion parameter values requires the robot agent to exploit its knowledge through reasoning. Some of the knowledge will be based on previous experience, some knowledge will be based on prospection using internal simulation, and some knowledge will be based on the robot's current perceptions. Note that the arguments of the query are exactly those values in the key-value pairs that were not passed as arguments to the generalized action plan, i.e. \textsf{\footnotesize ?object-to-be-fetched} and \textsf{\footnotesize ?destination}. Note also that the argument of the \textsf{\footnotesize to-succeed} function is exactly the body of the \textsf{\footnotesize Fetch\&Place} generalized action plan: that is, it forms a set of constraints on the values that are to be returned in the answer to the query by the KR\&R system.

  The contextualization of underdetermined action descriptions into
  concrete body motions requires KR\&R methods that can combine both
  abstract and concrete reasoning. Abstract reasoning enables open
  question answering by composing and chaining generalized and modular
  knowledge pieces. For example, through reasoning abstractly, robots
  can combine the knowledge that open filled containers should be held
  upright to avoid spilling with the knowledge that a milk carton is a
  container filled with milk and the knowledge that the milk carton is
  open, to conclude that the milk carton should be held upright when
  carried to the table. The power of generalized abstract knowledge is
  that it applies to all containers, even the ones that the robot does
  not know yet. At the same time the robot needs to make concrete
  inferences to decide on the body motion. It has to pick the hand
  shape for picking up the milk carton, the grasp points, and the
  grasp force. This requires to reason about the shape of the milk
  carton, its parts, its stability, the surface friction, the
  articulation model for opening and closing the milk carton, etc. In
  addition, contextualizing underdetermined action descriptions often
  requires predictive inferences, e.g. to assess the risk of the object
  falling out of the hand when grasping it in a particular way, taking
  into consideration whether the object has to be re-grasped before
  placing it at the intended location, or for assessing alternative
  body motions for accomplishing the action. To cover this range of
  inference tasks, CRAM employs digital twin knowledge representation
  and reasoning (DTKR\&R) which we will discuss in Section~\ref{sec:knowrob_in_CRAM}.

\subsection{The Generative Model}
\label{section:generative-model}
As we have said, CRAM uses a generative model to map from desired outcomes of an action to  motion parameter values. Conceptually, a generative model can be viewed as a joint distribution of  two sets of variables: the set of motion plan parameters generated by a generalized action plan, and the set  of physical effects (and, in particular, the robot's observations of the physical effects) that are the outcome of these motions. Thus, an underdetermined action description can be viewed as a request to sample a  joint probability distribution of motions generated by the associated generalized action plan  and the physical effects these motions cause. The contextualization query effectively samples this joint probability distribution to identify the motion parameter values that are most likely to succeed in accomplishing the desired outcome for the underdetermined action description. 
In CRAM, the generative model is realized, not yet as a probability distribution, but through knowledge representation \& reasoning, based on a robot's tightly-coupled symbolic and sub-symbolic knowledge representation, the tasks it is performing, the objects it is acting on,  and the environment in which it is operating.

The creation of the CRAM generative model was informed by interpretation and abstraction of high-volume and high-dimensional human everyday activity data. The priority was to ensure that the model can generate all fetch and place actions for setting the table, cleaning the table, loading and unloading the dishwasher, in realistic scenarios. These data include 100 multi-modal recordings of varied table setting episodes that contain muscle activity, eye tracking, motion capture, video, audio, concurrent and retroactive think-alouds as well as brain activity from mirror perception. These episode recordings are transformed into a machine-understandable representation, semantically rooted in a common ontology of robot and human agency and common representation of activities. 
 
The representation of  these episodes are referred to as  NEEMs (Narrative-Enabled  Episodic Memories).  NEEMs consist of a NEEM
 experience, which is a detailed, low-level, and agent-specific   recording of how the activity in the episode evolves, enriched with
 a NEEM narrative that provides information  explaining what is happening in the NEEM experience. They contain the  motion parameters and the observed effects of the motions. Robots  process NEEMs in order to abstract away from specific episode  contexts and learn the generally applicable common sense and naive physics knowledge that is part of the generative model. We say more about the representation and role of NEEMs in Section \ref{sec:knowrob_in_CRAM}.

The performance of robot agents can be substantially improved by adding knowledge to the generative model as well as extending the reasoning mechanisms and learning. Recall again that the underdetermined action descriptions in the  generative model are conceptualized as probability distributions over  fully-determined action instances. Thus, the expected performance of the generative model depends on the likelihood of the success of an action instance drawn from the distribution for accomplishing the respective request.

Even without knowledge, the generative model has a chance of achieving the respective request because it samples in the space of possible motion parameterizations. However, motion parameterizations that achieve the requests are extremely sparse. In addition, there is a high probability that the request can become unachievable because of unwanted side effects such as objects being broken or pushed out of reach.  As a consequence, this approach is not effective. We call this brute-force search generative model the uninformed model (\emph{gm\textsubscript{uninformed}}). 

Kazhoyan et al. \cite{kazhoyan17designators} have proposed a knowledge base with hand-coded heuristic rules encoding commonsense, intuitive physics, and other useful knowledge. Examples of heuristic rules are that in order to position yourself to detect an object you have to select a position from where the object is visible or you should first look for objects at places where you believe them to be. This knowledge base has proven to be effective but causes a substantial amount of backtracking.\footnote{The backtracking behavior is caused by the robot retrying to perform an action after its execution fails. The most common reason for this is the robot not being able to generate a collision-free trajectory for its arm to reach the goal. In that case the robot repositions its base and retries the action. The more confined is the space, where the object is located, the more challenging it is to find a collision-free trajectory for reaching it. For example, in the case where the action parameters, including the poses for the robot base, are inferred from the heuristics-based generative model, the robot repositions itself on average 3.6 times when grasping a milk box out of the fridge.} A positive aspect of the heuristic rules included in the knowledge base is that many of the rules are applicable to other kinds of actions such as pouring, wiping, or cutting, too. We call the generative model using the knowledge base consisting of heuristic rules the \emph{elementary proficiency level} model (\emph{gm\textsubscript{epl}}) and use it as the baseline performance for other variants of the generative model. 

We have implemented four extensions of \emph{gm\textsubscript{epl}} 
by augmenting it with 
(i)~prospective capablities 
\emph{gm\textsubscript{prospective}}, 
(ii)~experience-based learning capabilities 
\emph{gm\textsubscript{experience}},
(iii)~learning from observation capabilities
\emph{gm\textsubscript{imitation}}, and 
(iv)~transformational learning and planning capabilities
\emph{gm\textsubscript{transform}}.

The \emph{gm\textsubscript{prospective}} uses  a very fast simulation mechanism, which we call temporal projection of the intended action plan,  as an additional resource for selecting a promising motion parameterization \cite{kazhoyan19projection}. Using plan projection the robot agent can, for example, predict whether an intended grasp also allows the robot agent to place the object at the intended location. Projection reduces backtracking but can also cause delays in action execution due to the time required to run multiple instances of the simulation before making a decision. 

The \emph{gm\textsubscript{experience}} generative model records all queries for motion parameterizations, the returned parameterization, and whether the parameterized motion was successful in achieving the action goal \cite{koralewski19specialization}. These experience data are then used to learn how to parameterize the motions to maximize the  probability of success and the expected performance. Experience-based learning has the effect that the probability distributions that implement underdetermined action descriptions become more peaked and narrower. This means that the information content in the distributions is substantially increased and thereby the performance of \emph{gm\textsubscript{experience}} over \emph{gm\textsubscript{epl}} is significantly improved.

The \emph{gm\textsubscript{imitation}} generative model aims at performing the same learning tasks as \emph{gm\textsubscript{experience}} but creating the training data by observing humans operating  in a  virtual reality environment rather  than
having the robot collect its own  physical experiences \cite{kazhoyan20vr}.  The model learns much faster because humans
use their commonsense and intuitive physics knowledge when they generate training data.  A complication is that humans are much more dexterous than robots and the solutions have to be transferred from the human body  to the robot body to become executable. Again, we showed that \emph{gm\textsubscript{imitation}} achieves a significant performance improvement over \emph{gm\textsubscript{epl}}.

The \emph{gm\textsubscript{transform}} model is able to transform the motion plan schema by reordering motion phases and replacing specific motion phases with different ones \cite{kazhoyan20transformation}. These plan transformations provide
different ways of achieving behavior goals such as closing a drawer by pushing it with the elbow or a door with the foot instead of using the hand. Plan transformation sometimes opens up new optimization possibilities or makes actions
achievable by applying alternative strategies.


The knowledge that the robot agent acquires in order to improve its capability in accomplishing everyday manipulation tasks includes (a) the factorization  of the possible contexts into categories of contexts that require specific behavior patterns, (b) the optimization of the behaviors by tailoring them to the respective contexts, and  (c) the factorization and generalization of knowledge such that it is
composable and applicable to novel tasks and contexts.
 
\setlength{\tabcolsep}{0.3em} 

\begin{table}[t]
\caption{The ability to accomplish everyday activities can be improved by  adding prospection,  acquiring  experience,   imitation learning, and  plan transformation.  The baseline performance is given by   \emph{gm\textsubscript{epl}},  the elementary proficiency level. }
\vspace{-2mm}
\begin{tabular}{|l  p{0.24\textwidth}  p{0.08\textwidth}  p{0.05\textwidth} |}
\hline
\textbf{\small Model} & \textbf{\small Comments}  & \textbf{\small Reference \&  Dataset} & \textbf{\small Media }  \\ 
 \hline
  \emph{\small gm\textsubscript{uninformed}} & {\small Theoretically works but not effective.} &   & \\%

  \emph{\small gm\textsubscript{epl}} &  {\small Works effectively but needs substantial backtracking.}
    & {\small\cite{kazhoyan17designators} }&   {\small Video}\tablefootnote{\url{http://ease-crc.org/link/video-action-descriptions}}\\
  \emph{\small gm\textsubscript{prospective}} & {\small Significantly better than \emph{gm\textsubscript{epl}}.}
                                              \newline {\small Might delay execution.}
    &  {\small \cite{kazhoyan19projection}} & {\small Video\tablefootnote{\url{http://ease-crc.org/link/video-prospection}}  }\\ 

  \emph{\small gm\textsubscript{experience}} &  {\small Significantly better than \emph{gm\textsubscript{epl}}.}
                                                                 \newline  {\small Major resources required for experience acquisition.}
    &  {\small \cite{koralewski19specialization}}
        {\small Dataset\tablefootnote{\url{https://neemgit.informatik.uni-bremen.de/raw/iros-2019-plan-specialization}}} 
    &  {\small Video}\tablefootnote{\url{http://ease-crc.org/link/video-plan-specialization}} \\ 
  \emph{\small gm\textsubscript{imitation}} &  {\small Significantly better than \emph{gm\textsubscript{epl}}.}
                                                              \newline  {\small Relatively minor resources for experience acquisition }
                                                               {\small but requires adaptation to robot body.}
    &  {\small \cite{kazhoyan20vr}} 
        {\small Dataset}\tablefootnote{\url{https://neemgit.informatik.uni-bremen.de/raw/iros-2020-imitation-learning}}
    & {\small  Video}\tablefootnote{\url{http://ease-crc.org/link/video-imitation-from-vr}}\\
  \emph{\small gm\textsubscript{transform}} &  {\small  Tailors behaviors to specific contexts.}
                                             {\small  Typically little generality but high gain.}
    &  {\small \cite{kazhoyan20transformation}} &  {\small Video}\tablefootnote{\url{http://ease-crc.org/link/video-plan-transformations}}\\  
\hline
\end{tabular}
\vspace{-5mm}
\label{table:plan-based-control}
\end{table}

The results on generative models including references and video demonstrations are summarized in Table~\ref{table:plan-based-control}.  We are not aware of any other research initiative that demonstrates such a general, flexible, and context-guided 
accomplishment of human-scale manipulation tasks under such realistic circumstances. The respective experiments and their results are accessible as open research, including the open-source plans and the knowledge bases\footnote{\url{https://github.com/cram2/cram}} as well as the videos of the experiments and complete recordings of the experimental data  (see the links in the footnotes of Table~\ref{table:plan-based-control}).

Let us now proceed to describe how these four elements of the CRAM model of robot agency --- generalized action plan, motion plan, contextualization, and generative model --- are realized in the CRAM cognitive architecture.

\section{The CRAM Cognitive Architecture: \\A Realization of the Generative Model and Generalized Action Plan Framework}
\label{sec:CRAM_CA}

\begin{figure*}[t]
  \begin{center}
    \includegraphics[angle=0,width=0.75\textwidth] {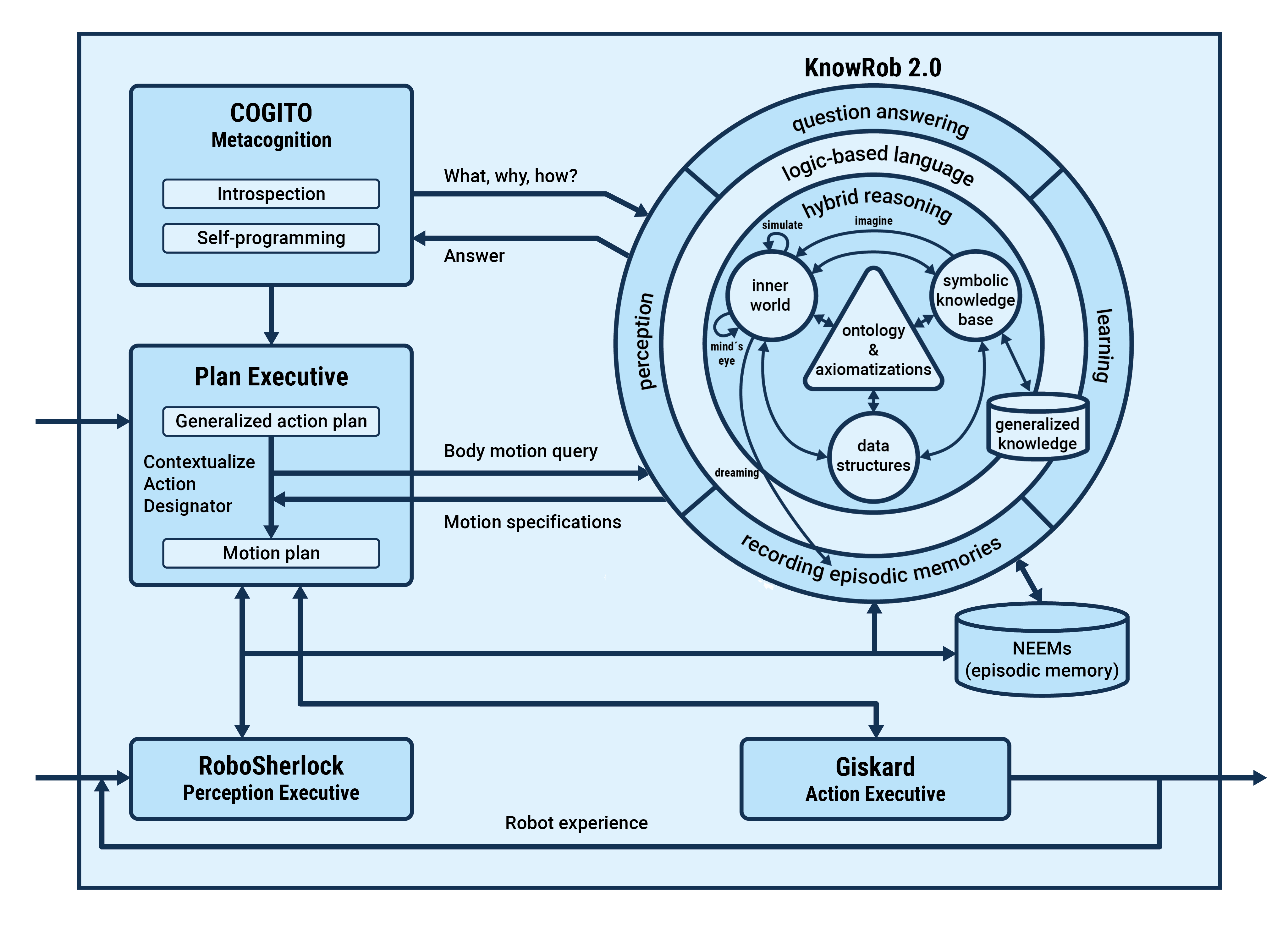}
    \caption{A schematic representation of the CRAM cognitive architecture with its five main components: the plan executive, the KnowRob 2.0 KR\&R system, the  metacognition system, the perception executive, and the action executive.  In KnowRob 2.0, the NEEMs (episodic memory) knowledge base and generalized knowledge base together comprise the episodic memory knowledge base (see text for details).}
    \label{fig:cram-cognitive-architecture}
  \end{center}
\end{figure*}
 
The four elements of the CRAM approach to carrying out everyday manipulation tasks described in the previous section are encapsulated in the CRAM cognitive architecture.\footnote{The CRAM open-source software is available at \url{http://www.cram-system.org.}}  It comprises five main components: (i) the CRAM Plan Language (CPL) executive; (ii)  a suite of knowledge bases and associated reasoning mechanisms, collectively referred to as KnowRob~2.0 \cite{Beetzetal18}; (iii) a perception executive referred to as RoboSherlock; (iv) an action executive referred to as Giskard; and (v) a metacognition system referred to as COGITO; see Fig. \ref{fig:cram-cognitive-architecture}.    In the following, we will discribe each component in turn.

\subsection{The CRAM Plan Language Executive}

The CRAM plan language executive --- plan executive, for short ---  interprets and executes generalized action plans written in the CRAM plan language (CPL). It does so by determining the difference between what knowledge is required by the motion plan and what is known about the action so far in order to formulate a query for the  motion parameter values to fill these gaps in the required knowledge.  The query is then answered by sampling the generative model, using knowledge embedded in the plan,   the KR\&R system KnowRob~2.0, and  perceptual information provided by the perception executive RoboSherlock. The  resultant motion plans are then executed by the action executive Giskard.  

For a comprehensive treatment of CPL, see \cite{Mosenlechner16}. Here, we will provide a sample of the key constructs of the language and then look again at the \textsf{\footnotesize fetch\&place} generalized action plan.
 
CPL is an extension of Lisp, exploiting  macros and functions to create a task-specific language for programming cognitive robots. It  makes heavy use of multi-threading, especially for fluents, one of the key
extensions. A fluent is a proxy object for some Common Lisp object, typically interfacing to some sensor, and it is used as a variable that allows a separate thread to  monitor and act on a change in the value of that object using  functions such as \textsf{\footnotesize (wait-for fluent)}, which blocks until a value is available, or \textsf{\footnotesize (wait-for pulsed fluent)}, which blocks until the value changes. Different handling options are available to cater for fluent values that are pulsed while a previously pulsed value is being processed, effectively buffering one or all pulsed values, or none. 

CPL supports concurrency, with various sequential execution constructs (e.g. \textsf{\footnotesize seq}, \textsf{\footnotesize try-in-order}) and parallel execution constructs (e.g.  \textsf{\footnotesize par}, \textsf{\footnotesize pursue}, \textsf{\footnotesize try-all}), each offering different ways to advance the thread of execution. Failure in executing actions is a normal occurrence in robotics, especially when carrying out everyday activities,  so CPL allows failures to be thrown and caught using the \textsf{\footnotesize handle-failure} function. This function specifies the failure-handling code and provides control over the number of attempts to re-try the code that threw the failure, after first having taken some remedial action. It also has a full-featured Prolog interpreter, written in Lisp, to provide for the definition of facts and predicates, and for reasoning.  

We have already met one of the main constructs in CPL: the designator.  To the robot plan programmer, designators are objects containing sequences of key-value pairs of symbols, the second value element of each pair acting as a placeholder for information that is required by the CRAM plan or a placeholder for the execution of a motor command.  The information is acquired or the motor command executed by resolving the designator at run time in the current context of the task action. Information is acquired by querying a priori knowledge embedded in the plan, by querying knowledge in the KR\&R  KnowRob~2.0 knowledge base, and by accessing sensor data through the perception executive, either directly or via a feature in KnowRob~2.0 called a computable predicate, about which we will say more in Section \ref{sec:knowrob_in_CRAM}.  

As we mentioned in Section \ref{sec:ease-approach}, there are four types of designator: motion designators (e.g. for motor commands), location designators (e.g. for 3D poses), object designators (e.g. for grasp configurations), and action designators (e.g. for high-level goal-directed functionality).  Each type of designator has a different resolution strategy. For example, motion designators are resolved by the action executive. Object designators provide a direct interface between CRAM plans and the RoboSherlock perception executive. The key-value pairs in the designator's properties describe the object that is to be perceived and  RoboSherlock  resolves the designator by extracting the required information from the sensor data. Location designators are typically resolved as robot poses that are appropriate for manipulating an object.  Resolution is accomplished in two steps: (i) generation of a lazy list of candidate poses, and (ii) testing that candidate poses are feasible. This facilitates a general generation process and a specific filter process to remove the invalid solutions.  Action designators are usually resolved using the Prolog inference engine to convert the high-level symbolic action into constitutent lower-level action designators,  motion designators, location designators, and object designators, all of which are then resolved accordingly. 

A CRAM plan, in general, and a generalized action plan, in particular, comprises one or more action designators embedded in CPL (and Lisp) control structures.  We have already described the concept of a generalized action plan in Section \ref{section:generalized-action-plans}, including an outline of the \textsf{\footnotesize fetch\&place}  generalized action plan.  One of the key features of CRAM is that the \textsf{\footnotesize fetch\&place}  generalized action plan  can generate the context-specific behavior required to accomplish everyday activities, such as  setting the table for a meal and tidying up afterwards.   The distinctive feature of the plan is the variability of fetch and place behavior that a single plan schema can generate. These variations include active object search, optional support actions such as opening and closing containers in order to fetch objects, and  the coordination of bimanual pickup actions. It also provides for  dynamic behavior adaptation, including  skipping of unnecessary subactions. In addition, it includes sophisticated methods for failure detection and handling,  as well as the context-dependent continuation of the primary activity after the failure recovery.

The plan executive  uses  explicit symbolic knowledge structures to represent the behavior-generating plans, the computational processes they initiate, the motions they generate, and the physical effects that the motions cause \cite{mosenlechner2010logs}. It also represents the causal relationships between these knowledge structures. This  knowledge  enables the robot agents to answer queries about what the robot does, why it does it, how it does it, and
what is happening. It also allows the robot agent to diagnose its behavior by inferring answers to questions such as: ``Is the goal of the action achievable?'' and ``Did the action fail because the robot did not see the object?''  These knowledge structures allow  the robot to identify the subplans that are responsible for certain effects. This enables powerful plan transformation, achieved by enhancing the generative model, as described in Section \ref{section:generative-model}.

\subsection{KnowRob~2.0: The Knowledge Representation \& Reasoning Framework}
\label{sec:knowrob_in_CRAM}

KnowRob 2.0  is CRAM's knowledge representation \& reasoning framework. KnowRob~2.0 enables robots to acquire open-ended manipulation skills, reason about how to perform manipulation actions, and acquire commonsense knowledge. It has enabled robot agents to accomplish complex manipulation tasks such as making pizza, conducting chemical experiments, and setting tables.  

KnowRob 2.0 is implemented in Prolog and it is exposed as a conventional first-order time interval logic knowledge base. However, many logic expressions are constructed on-demand from sensorimotor data computed in real-time, through  algorithms for motion planning and  inverse kinematics, for instance, and from log data stored in noSQL databases.   It provides the background common sense intuitive-physics knowledge required by the plan executive to implement its goal-directed under-determined task plans, e.g. how to grasp an object, depending on the object’s shape, weight, softness, and other properties; how it has to be held while moving it, e.g. upright to avoid spilling its contents; and where the object is normally located.  Some knowledge is specified {\em a priori}, some is derived from experience, and some is the result of simulated execution of candidate actions using a high-fidelity virtual reality physics engine simulator. All knowledge is represented by a first-order time interval logic expression, and reasoned about, as needed. 

KnowRob 2.0 comprises five core elements embedded in a hybrid (i.e. multi-formalism) reasoning shell, exposed through a logic-based language layer to an interface shell that provides functionality for perception, question answering, experience acquisition, and knowledge learning; see Fig. \ref{fig:cram-cognitive-architecture}. The five elements are: (i) a central set of knowledge ontologies and axiomatizations; (ii) an episodic memory knowledge base encapsulating the robot's experiences, comprising the NEEMs symbolic \& sub-symbolic knowledge base and a generalized symbolic knowledge base;  (iii) an inner world knowledge base and virtual reality physics engine simulator; (iv) a symbolic knowledge base with abstracted symbolic sensor and action data, logical axioms, and inference rules; and (v) a virtual knowledge base comprising a set of data-structures for parameterized motion control and path planning.

\subsubsection{Knowledge Ontologies and Axiomatizations}

CRAM employs a collection of ontologies \cite{Bateman-etal2018-ease} within a  concise foundational or top-level ontology, called SOMA (socio-physical model of activities). SOMA is a parsimonious extension of DUL (Dolce Upper Lite) ontology, where additional concepts and relations provide  deeper semantics of autonomous activities, objects, agents, and environments. SOMA has been complemented with various subontologies that provide background knowledge on everyday activity and robot and human activity including axiomatizations of NEEMs,  common models of actions, robots, affordances \cite{bessler20affordances},  and execution failures \cite{diab2019ontology}.

\begin{figure}[t]
  \centering
  \includegraphics[width=1.0\columnwidth]%
  {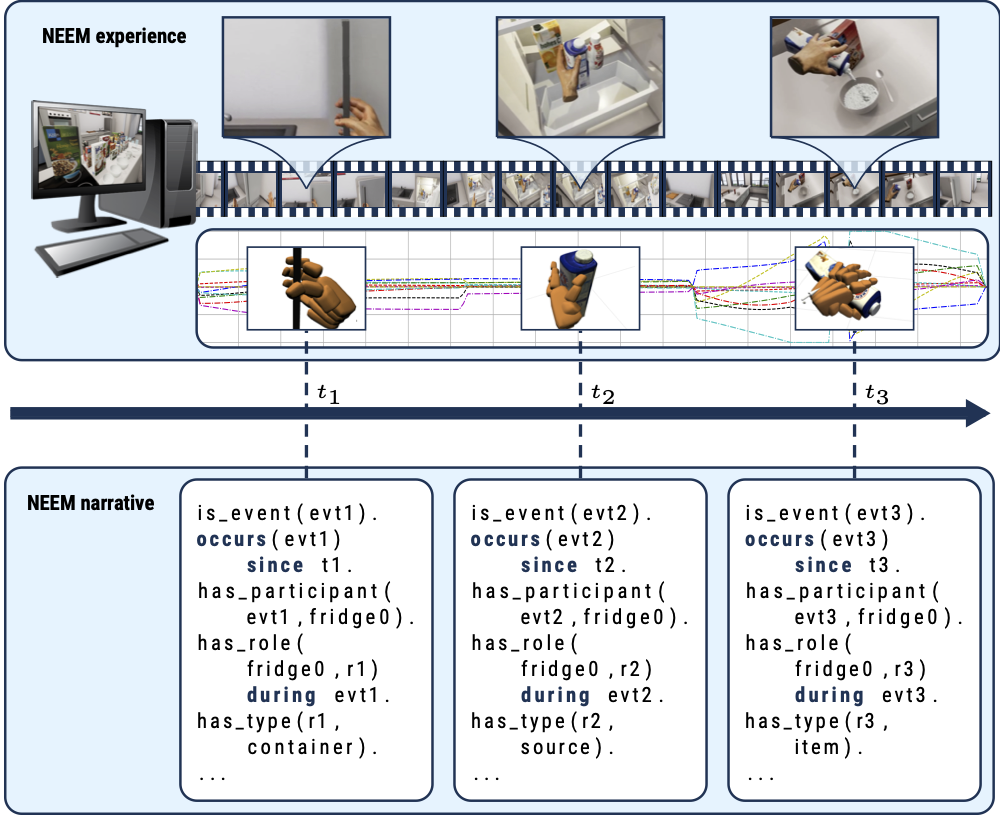}
  \caption{Schematic visualization of a NEEM: sub-symbolic experiential episodic data and motor control procedural data, augmented by a symbolic  description of the episode as it unfolds.}
  \label{fig:neem-representation}
\end{figure}

\subsubsection{Episodic Memory Knowledge Base}
One of the  distinguishing aspects of KnowRob 2.0 is its focus on episodic memory. The episodic memory knowledge base comprises the NEEMs knowledge base and a generalized knowledge base; see Fig. \ref{fig:cram-cognitive-architecture}.    NEEMs, introduced above in Section \ref{section:generative-model}, are inspired by models of the human episodic memory system, which refers to a type of declarative memory that contains autobiographical events. When an episodic memory is recalled, it results in the retrieval of the whole context of the relevant episode, including sensory, affective and cognitive processes. Semantic information such as general facts and concepts are thought to be derived from accumulated episodic memory \cite{Tulving72}. NEEMs are a way of storing the data generated by robot agents during everyday manipulation in such a way that enables knowledge extraction. More formally, NEEMs are CRAM's generalization of episodic memory, encapsulating sub-symbolic experiential episodic data, motor control procedural data,  descriptive semantic annotation, and the accompanying mechanisms for acquiring them and learning from them in KnowRob~2.0.  NEEMs are an agent's memories   of activities that it executed, observed, simulated, or read about. A NEEM of  an activity consists of the NEEM experience, which is a   detailed, low-level, and agent-specific recording of how the activity in the episode evolves, enriched with the NEEM narrative, which is a story providing information that  explains what is happening in the NEEM experience; see Fig. \ref{fig:neem-representation}. Agents  collect and store NEEMs in their NEEM system and process them in order to abstract away from specific episode contexts and learn the   generally applicable commonsense and naive physics knowledge needed   for mastering everyday activities.

While performing an activity, such as setting a table, the robot logs its perception and execution data in great detail. This includes sensory data (images, body poses, etc.)  and control signals.  These records of external perceptions and the internal semantically annotated control structures enable the robot to look at the low-level data as if they were virtual stories --- narratives --- about performing the activity in different ways, where the robot's intentions, beliefs and behavior, perceived scenes, and effects of actions are related to each other.  Because they are framed with respect to concrete episodes rather than  generalized knowledge, this story view enables the robot to answer questions regarding to what it did, why, how, how well, etc.  The robot can answer queries such as: ``Where do I find clean cups?'', ``Which is the best order to bring items to the table?'', ``At what times should the table be set?'', and ``Which perception routines work best for detecting plates in the cupboard?''

NEEMs enable the agent to replay specific experiences with its \textsl{mind's eye} and, for example, recall meaningful sub-episodes of successfully picking up a red cup.  The agent can use these past episodes to learn new information, even for aspects that were not previously considered for that particular episode.

\begin{figure}[tb]
  \begin{center}
    \includegraphics[width=\columnwidth] {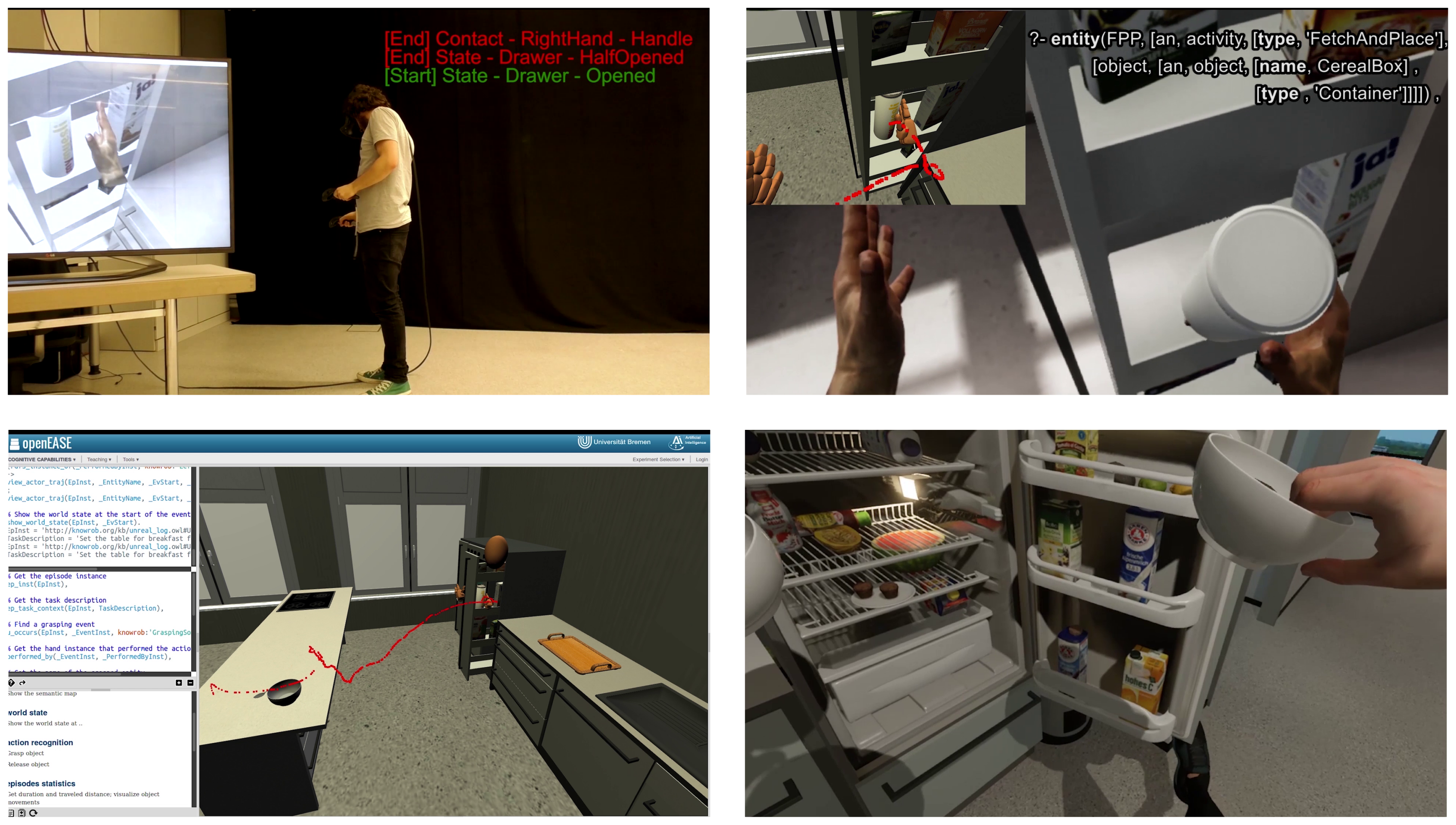}
    \caption{Collecting NEEMs from human demonstration in a virtual kitchen environment.}
    \label{fig:blackbox-ameva}
  \end{center}
\end{figure}

KnowRob~2.0 provides a query language in order to retrieve information from NEEMs. The expressive power  --- that is, the set of questions that can be asked about a given NEEM --- is provided by the KnowRob~2.0 ontology. One can retrieve all entities of a given entity category in the KnowRob~2.0 ontology and describe each entity using the attributes defined for the respective entity category. In addition, the relations defined in the ontology can be used to constrain combinations of entities.  The assertions about entities and relations are automatically generated from the EASE ontology.  The integration of the NEEM database, referred to as the NEEM Hub, with the SOMA ontology in KnowRob 2.0 yields a powerful hybrid symbolic / sub-symbolic framework for observation and interpretation of activities for reasoning and forms the basis of the CRAM generative model.  We noted in Section \ref{section:generative-model} that the generative model was created using data gathered from human activity. This data is encapsulated in NEEMS generated by humans, both as a result of physical actions carried out by sensorized humans in the real world and as a result of actions carried out by humans operating  in a high-fidelity photorealistic virtual reality environment. The NEEMs were captured using AMEvA (Automated Models of Everyday Activities) \cite{haidu19ameva}, a computer system that can observe, interpret, and record fetch-and-place tasks and automatically abstract the observed activities into action models \cite{Flanaganetal2006}, and then represent these models as NEEMs.  Fig.~\ref{fig:blackbox-ameva} shows the operation of AMEvA in which a human is performing \textsf{\footnotesize fetch\&place} tasks in a virtual kitchen environment. 

The usefulness of NEEMs recorded by robot agents has also been demonstrated in the experience-based learning generative model
\emph{gm\textsubscript{experience}} discussed in Section~\ref{section:generative-model}.

NEEMs are also available at execution time. This means that the robot agent can use the active NEEM to diagnose and recover from execution failures, given that the last time instant of the active NEEM is the current belief state of the robot agent
\cite{bartels2019episodic}.

\begin{figure}[t]
  \centering
  \includegraphics[width=\columnwidth]  {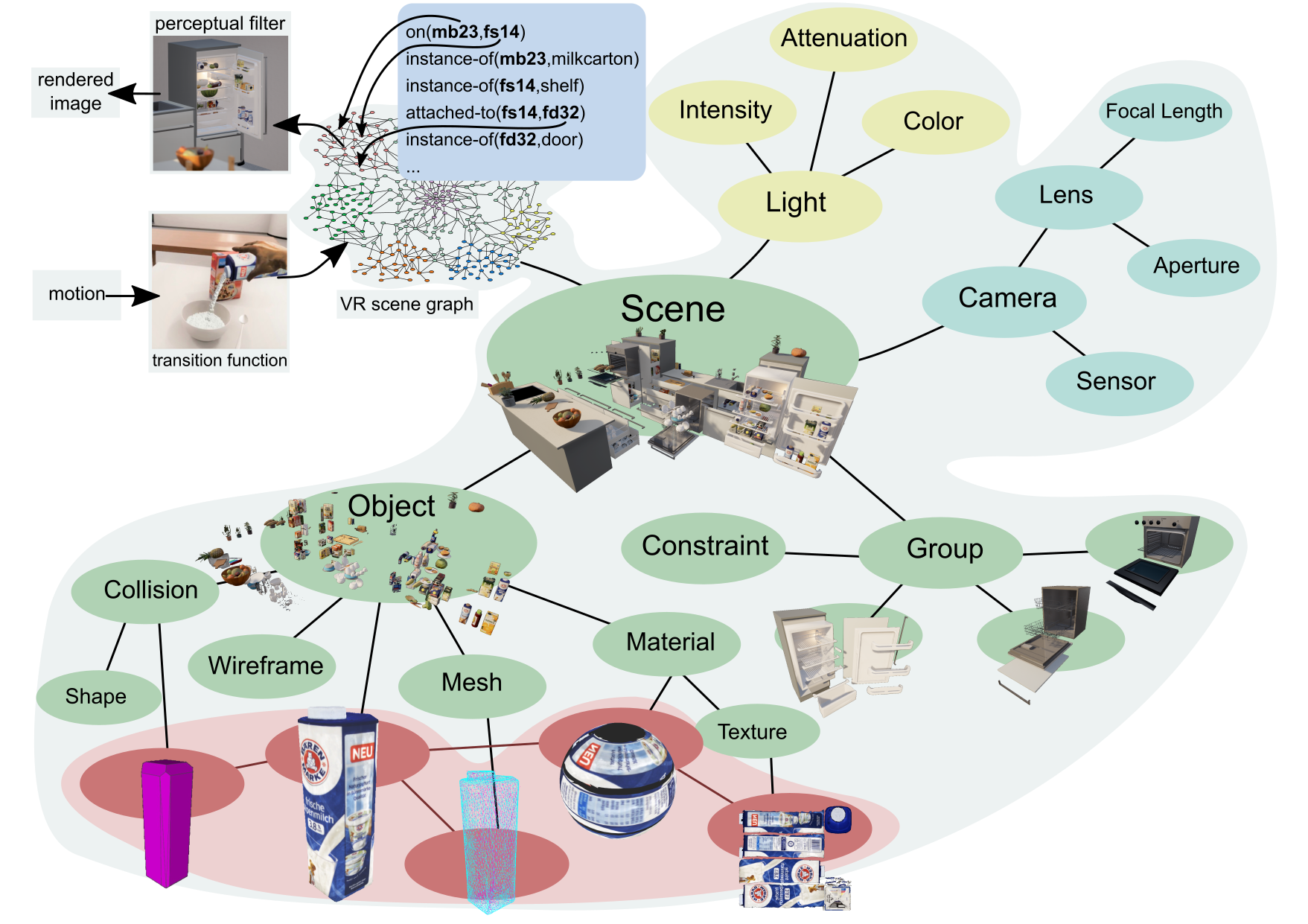}
  \caption{Symbolic knowledge in KnowRob is grounded in a virtual reality scene graph representation (top left of the diagram) which, in turn, provides the basis for digital twin knowledge representation and reasoning (exploded view of part of the scene graph in the center, bottom, and right of the diagram).}
  \label{fig:dtkrr}
\end{figure}

\subsubsection{Inner World Knowledge Base}

\begin{figure}[t]
  \centering
     \includegraphics[width=1.0\columnwidth]%
     {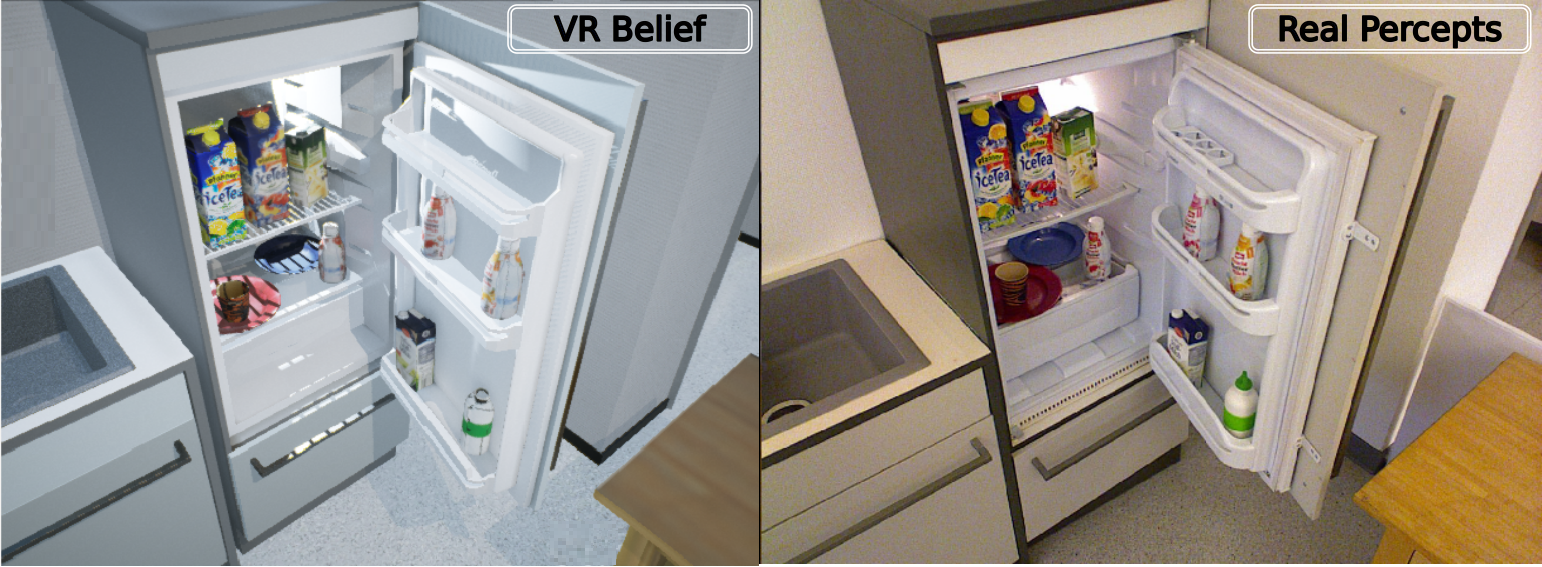}
    \caption[Expectation for belief state estimation in
    RoboSherlock]%
    {Expectation for belief state estimation generated through  imagistic reasoning. The belief image on the left is a  rendering of the symbolic belief state of the robot agent.}
    \label{fig:imagistic-reasoning}
\end{figure}

The  inner world knowledge base, also called a digital twin knowledge representation and reasoning system (DTKR\&R), facilitates geometric reasoning using a photo-realistic virtual reality system and physics engine.   The DTKR\&R uses the scene graph data structures that are the implementation basis of  virtual environments and integrates them with the symbolic knowledge representation system, providing every entity that is relevant for the robot agent, be it an object, object part, an articulation model, or a sub-scene, with a symbolic name; see Fig. \ref{fig:dtkrr}. This symbolic name is then axiomatized in the symbolic knowledge base by asserting it as an instance of an object category defined in the ontological knowledge base and providing formally-stated background knowledge about the entity. The relationship between the symbolic knowledge base  and the inner world means that every relevant physical entity of the artifical world is formalized in the knowledge base and every symbolic physical entity is also an entity in the inner world. 
This means that the symbolic representations are grounded, not in physical objects, but in their counterparts in the inner digital twin world. However, this inner world is continually synchronized with the real world though perceptual data provided by the perception executive (to be described in the next section). In this way, the DTK\&R inner world finesses the symbol grounding problem and provides the means by which the CRAM generative model represents the belief state of a robot agent, using DTKR\&R\ as an artificial world which can be visually rendered and physically simulated. Thus to assess the information content of the belief state at a time instant \emph{$t_i$} during an everyday activity episode we can capture an image from the artificial world at time instant \emph{$t_i$} and compare it with an image captured by a real camera in the real environment.  Figure~\ref{fig:imagistic-reasoning} shows an example of an image showing the belief state generated by this process and the comparison with the real state of the environment at the same time instant.

The DTKR\&R\ provides a comprehensive,  concrete belief state for the robot.  In terms of comprehensiveness, it
matters whether the belief state contains all relevant objects and whether the scene representations are sufficiently rich for
contextualizing object manipulation. The first criterion for sufficient accuracy is that the parameters for the motions that implement object manipulation tasks can be inferred sufficiently accurately so that the manipulation actions will succeed. As the belief state has a sub-symbolic rendering, which is very similar to the images captured by robot cameras, it supports the grounding of symbols in the perception-action loop. The second criterion concerns the accessibility of the knowledge in the belief state. In particular, it concerns the ability of robot agents to infer answers to the body motion queries based on the belief state. For example, in order to fetch a spoon in the context of setting a table, the robot should be able to infer answers from its belief state for questions including the following ones: ``Where can I find spoons that are clean and unused?'' ``How can the container, which the robot believes the spoons to be in, be opened and closed?'' ``How should the drawer be grasped to open it?'' ``What is the handle of the spoon by which it should be grasped?''

Knowledge in the belief state can be retrieved through asking queries. For example, the robot could ask queries such as: ``which
containers open counterclockwise?'', ``which objects are electrical  devices?'', or ``what is a storage space for perishable items?''. The refrigerator is an answer to all of these queries. If the robot then asks for knowledge about the refrigerator, the accessible knowledge would include the part hierarchy of the refrigerator, including the 3D models of the parts, the articulation model of the refrigerator that tells the robot how to open and close it, and so on. 

Note that in order to obtain this question answering capability we make the \emph{weak closed world assumption}; 
that is, in its reasoning processes the robot assumes to know all objects but concurrently it monitors its percepts for new objects and updates its belief state whenever a new object is detected.
The knowledge base
of the robot is populated with object models that consist of CAD models, including the part structure and possible articulation models, a texture model, as well as encyclopedic, commonsense, and intuitive physics knowledge about the object. We call the closed-world assumption \emph{weak} because the robot is still required to detect novel objects, for example if the robot unpacks a shopping bag or somebody else put a novel object in the environment. Under the \emph{weak closed world assumption}, computing the belief state comes down to maintaining a belief about where each object is and which state it has. Domain knowledge can be provided as prior knowledge through a hand-coded ontology, which contains valuable manipulation knowledge, such as a container can be opened by generating a motion implied by the articulation model of the container (e.g. the knowledge that a screw top cap can be removed by twisting the cap). Assuming a weak closed world, robot agents in CRAM can also answer queries that require prospective capabilities. An example of such a query is: ``what do I expect the inside of the refrigerator to look like when I open it?''  Answering this query requires the robot agent to visually render the scene inside the refrigerator given its current belief state. Another example is: ``what do I expect to happen if I pick up the object in front of me with my right gripper?''

Since the DTKR\&R allows KnowRob 2.0  to simulate the outcome of candidate action and to establish the feasibility of that action, the inner world knowledge base serves two roles: a representation of the belief state of the robot about itself and the world (as describe above), and a reasoning mechanism for determining the outcome of candidate actions. As such, the inner world knowledge base encapsulates two types of knowledge: current beliefs about robot and the world, and projected internal simulation of future states. It also acts as a learning mechanism, generating NEEMs off-line and running simulations of activities with varying control parameters.  These are recorded and transferred to the episodic memory knowledge base, i.e. the NEEM Hub.

\subsubsection{Symbolic Knowledge Base}

The symbolic knowledge base provides information about the entities in the robot’s environment, including objects, object parts, object articulation models, environments composed of objects, actions, and events. It uses an entity description language that allows partial description of entities in terms of both symbolic and sub-symbolic properties.

\subsubsection{Virtual Knowledge Base of Data Structures}
The virtual knowledge base provides computable predicates that facilitate the integration of non-symbolic data, e.g. perceptual information, into the reasoning process,  allowing symbolic queries of non-symbolic data.   This allows run-time sensorimotor states to be integrated into the knowledge base at run-time and to be used in reasoning in the same way as symbolic knowledge.

\subsubsection{Access to KnowRob 2.0}

As we noted at the beginning of this section, KnowRob 2.0  provides a logic-based language interface that allows the hybrid KR\&R system to be exposed as a purely symbolic knowledge base even though internally it uses multiple symbolic and sub-symbolic representations and reasoning formalisms.
In this way, KnowRob 2.0  can be treated by the plan executive (and other systems through its OpenEASE interface \cite{Beetzetal15}) as a symbolic object-oriented query system in which entities can be retrieved by providing partial descriptions of them using the entity predicate.  This allows KnowRob 2.0 to appear as a ``Siri for robots'', i.e. as a query and response oracle. Consequently, during task execution, there is an on-going dialogue between the plan executive and KnowRob 2.0, in which the plan executive presents a series of underdetermined queries for motion plan parameter values and KnowRob 2.0 provides the corresponding responses, i.e. the values that are likely to lead to a successful action, allowing the plan executive to carry out the task using the action executive.

\subsection{RoboSherlock: The Perception Executive}
\label{sec:perception}

The resolution  of object designators, providing information about the objects in the robot's visual field of view, is accomplished by  the RoboSherlock perception executive \cite{beetz15robosherlock}. RoboSherlock provides a symbolic language that enables the robot agent to specify a large variety of perception tasks that need to be performed in order to accomplish underdetermined everyday manipulation tasks. In this language, perception tasks are stated in terms of object   descriptions, object hypotheses, and task   descriptions.  Using these descriptions, a robot agent can describe a red spoon using the following construct: \textsf{\footnotesize (an object (category spoon) (color  red))}. The function \textsf{\footnotesize detect} applied to an object description causes the perception system to detect objects in the sensor data that satisfy the description and return the detected hypotheses.  An object detection task has the following form:

\singlespace
\vspace{-3mm}
\textsf{\footnotesize
\begin{tabbing}
~~~ \textsf{\textbf{detect}} (\textsf{\textbf{an}} \= \textsf{\textbf{object }}\\
                         \> (category ?category) \\
              \>                                 ($\langle$key-1$\rangle$ $\langle$value-1$\rangle$) \\
               \>                                 ($\langle$key-2$\rangle$ $\langle$value-2$\rangle$) \\
              \>                                 \ldots)
\end{tabbing}
}

\setlength{\tabcolsep}{0.2em} 

\begin{figure}[t]
  \centering
  \begin{tabular}{|p{0.41\columnwidth} p{0.55\columnwidth}|}
    \hline
    \textbf{\footnotesize Query Results} &
    \textbf{\footnotesize Perception Task} \\
    \hline
    \begin{minipage}{.41\columnwidth}
    \includegraphics[width=\columnwidth]{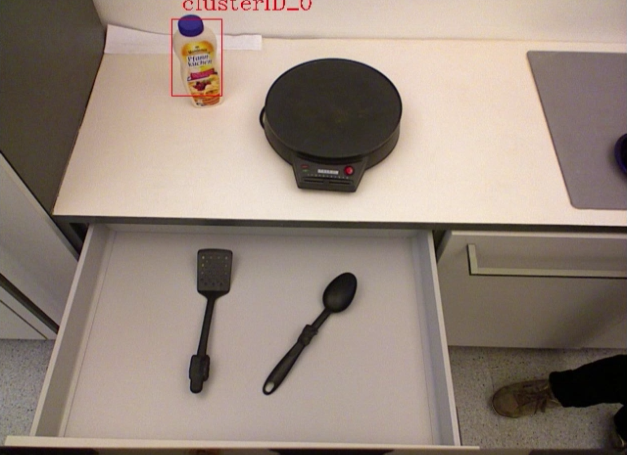}
    \end{minipage}
    & 
    \begin{minipage}{.55\columnwidth}
        \begin{tiny}
\textsf{
        \begin{tabbing}
    \textbf{detect} (an \= object \\
    \> (category `FoodOrDrinkOrIngredient'))\\
    \textbf{examine}\footnote{\scriptsize Examine tasks can only be called for existing object hypotheses} ?id [:size, :pose] \\
    \textbf{detect} (an \= object \\
    \> (color yellow) \\
    \> (shape round)) \\
    \textbf{detect} (an \= object \\
    \> (location (a \= location \\
    \>              \> (next-to (an \= object \\
    \>              \>               \> (category `ElecDevice'))))))\\
    \textbf{examine} ?id [:geom-primitive, :logo]
        \end{tabbing}
}
        \end{tiny}
    \end{minipage}
    \\
    \hline
    
    \begin{minipage}{.41\columnwidth}
    \includegraphics[width=\linewidth]{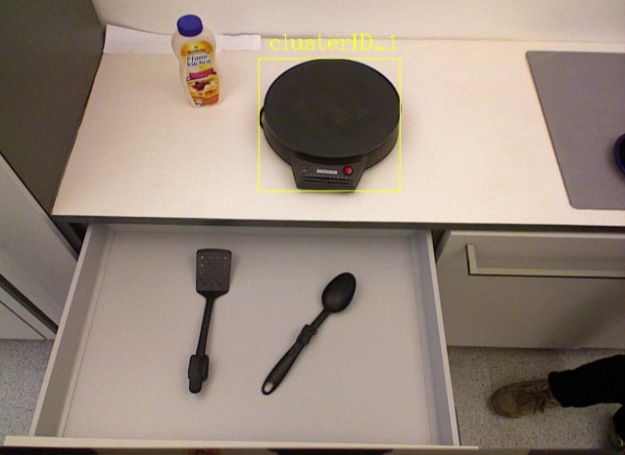}
    \end{minipage} 
    &
    \begin{minipage}{.55\columnwidth}
        \begin{tiny}
\textsf{
        \begin{tabbing}
    \textbf{detect} (an \= object \\
    \> (category `PancakeMaker') \\
    \> (location (a \= location \\
    \>              \> (on (an \= object \\
    \>              \>         \> (category `CounterTop')))))) \\
    \textbf{detect} (an \= object \\
    \> (color black) \\
    \> (shape round) \\
    \> (category 'ElecDevice'))
        \end{tabbing}
}
        \end{tiny}
        \end{minipage}\\

    \hline
    \begin{minipage}{.41\columnwidth}
    \includegraphics[width=\linewidth]{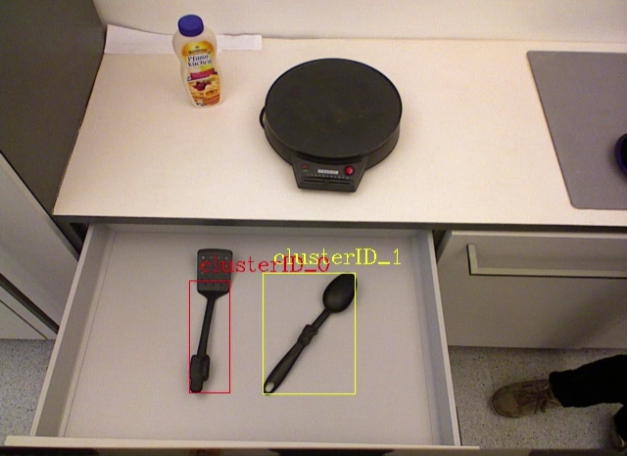}
    \end{minipage} 
    &
    \begin{minipage}{.55\columnwidth}
        \begin{tiny}
\textsf{
        \begin{tabbing}
    \textbf{detect} (an \= object\\
    \> (category `CookingUtensil') \\
    \> (location (a \=location \\
    \>              \>(in (an \= object \\
    \>              \>        \>(type container) \\
    \>              \>        \> (category drawer\#3))))))\\
    \textbf{detect} (an \= object\\
    \> (shape flat) \\
    \> (color black) \\
    \> (location (a \= location \\
    \>              \> (in (an \= object \\
    \>              \>         \>(type container) \\
    \>              \>         \> (category drawer\#3))))))
        \end{tabbing}
}
        \end{tiny}
        \end{minipage}\\
    \hline    
  \end{tabular}

  \caption{RoboSherlock as a vision-based question answering system.}
  \label{fig:robosherlock-blackbox}
\end{figure}

\begin{figure}[t]
  \centering
  \includegraphics[width=\columnwidth]%
  {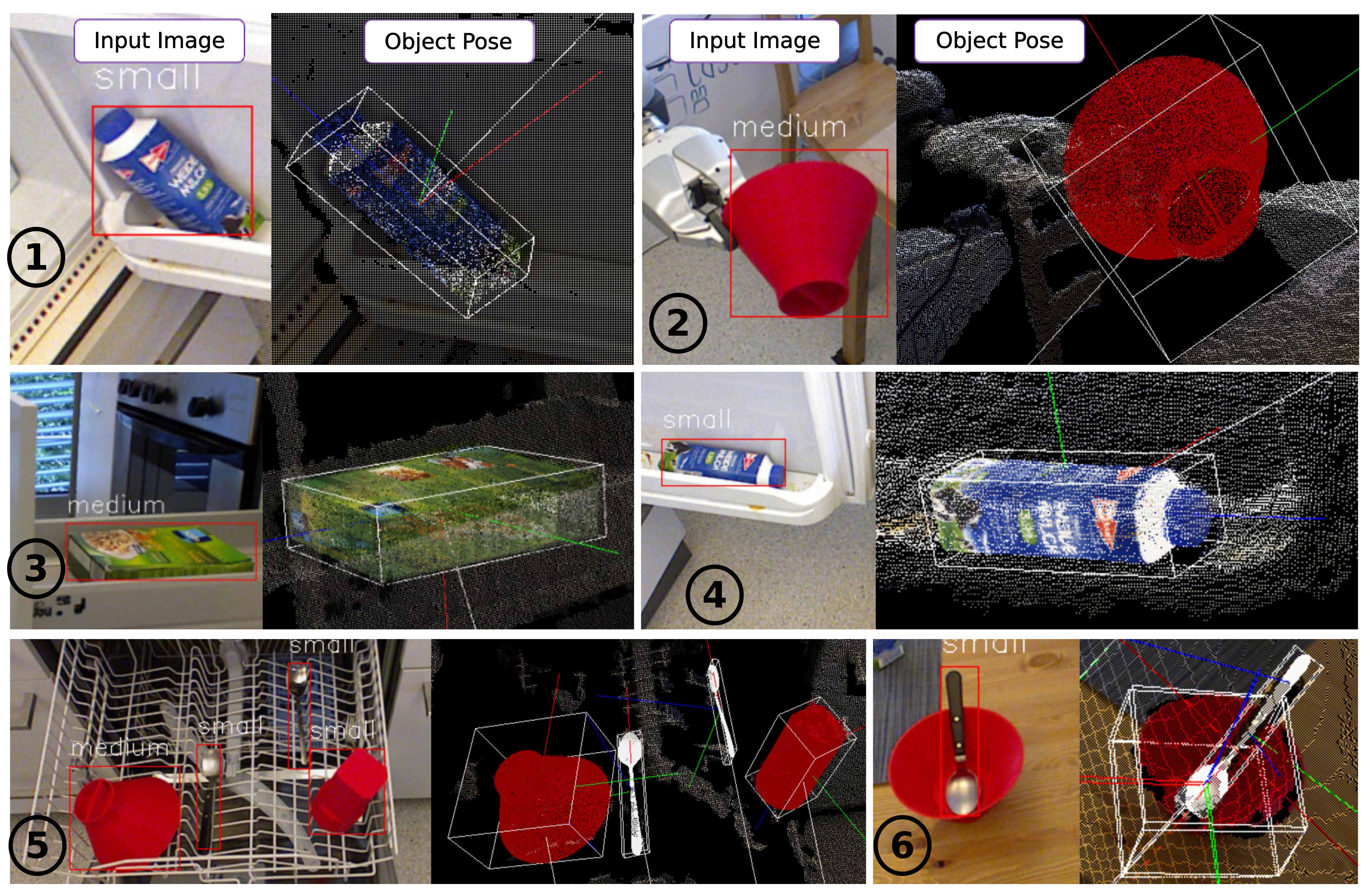}  
  \caption[Deep network-based object detection and object pose
    estimation in RoboSherlock]{Deep network-based object detection and object pose
    estimation in challenging configurations and context.}
  \label{fig:object-detection}
\end{figure}

The attributes that can be used in object descriptions include \textsf{\footnotesize shape}, \textsf{\footnotesize color}, \textsf{\footnotesize category}, \textsf{\footnotesize location}, \textsf{\footnotesize pose}, \textsf{\footnotesize CAD-model}, and \textsf{\footnotesize part-of}. In particular, the \textsf{\footnotesize category} attribute is very expressive as it allows for the application of self-defined categories. Given a classifier that can infer the affordances of object hypotheses,  the perception system can detect objects in a scene that afford a given action. Also, by combining visual detection with knowledge-enabled reasoning and other forms of computations, such as computing volumes, RoboSherlock can also accomplish perception tasks such as the detection of objects that satisfy some condition, e.g. ``a container that can hold more than half a liter''.
 
RoboSherlock uses an extensible ensemble of experts to accomplish perception tasks. RoboSherlock\ perception experts are special-purpose routines that are employed in the respective perception contexts. Thus, rather than applying a general-purpose plate detector, RoboSherlock uses context-specific plate detectors. During table setting, it might use one that detects the topmost white horizontal lines in cupboards, while it uses detectors for ovals when cleaning the table. It also might use texture detection when it assumes the plates to not be clean, and so on. Thus, given a perception task and the current context information, RoboSherlock executes the appropriate perception experts to generate candidate solutions for the given perception task and even generates context-specific perception pipelines. The candidate solutions proposed by experts are then tested and ranked to find the best solution. The advantage of RoboSherlock over other perception approaches is that it can combine knowledge with perception and use knowledge-enabled reasoning about objects and scenes to tailor perception capabilities to the respective contexts and thereby make it more effective, robust, and efficient.

Fig.~\ref{fig:robosherlock-blackbox} shows examples of the use of the \textsf{\footnotesize \textbf{detect}} construct. Fig. \ref{fig:object-detection} shows some of the current capabilities of RoboSherlock in more complex scences,  including a cluttered fridge, dishwasher, and oven.

RoboSherlock exploits the power of RobotVQA \cite{kenghagho20paperiros},  a scene-graph and deep-learning-based visual question answering system for robot manipulation. At the heart of \textsc{RobotVQA} lies a multi-task deep-learning model that infers formal semantic scene graphs from RGB(D) images of the scene at a rate of about 5 frames per second. The graph is made up of the set of scene objects, their description (category, color, shape, 6D pose, material, mask) and their spatial relations. Moreover, each of the facts in the graph is assigned a probability as a measure of uncertainty. The estimated scene graphs are represented using a probabilistic lightweight description logic.

RoboSherlock maintains an internal belief state implemented through virtual reality technologies; see Fig. \ref{fig:imagistic-reasoning}. The knowledge base of the robot is populated with object models that consist of CAD models, including the part structure and possible articulation models, a texture model, as well as encyclopedic, commonsense, and intuitive physics knowledge about the object. This imagination-based scene perception approach has the advantage that the robot has perfect knowledge about everything that is contained in the belief state. A second advantage is that the robot can compute  detailed and realistic image-level expectations about what it expects to see. These expectations are used to estimate object poses  accurately and to save computational resources by rendering the belief state  from the camera pose as an image and by comparing it with the image captured by the robot camera in the real environment.

RoboSherlock increasingly uses self-supervised learning methods leveraging these inner-world models of the robot
environments \cite{mania2019framework,kenghagho20paperiros}, which together with the images contained in the episodic memories (i.e. NEEMs) \cite{balint2018variations,balintbe17icar} are sufficient to learn robust real-world perception methods.
This framework enables robot agents to use their environment and object representations in order to generate training data for supervised learning for perception tasks. For training, the framework  not only allows the creation of typical scenes in the environment but also the generation of distributions for typical robot behaviors. This way, the distribution of training data can be tuned to specific kinds of environments and tasks.

\begin{figure*}[tp]
  
  \centering
  \begin{tabular}{c c c}

    \includegraphics[height=33ex]%
    {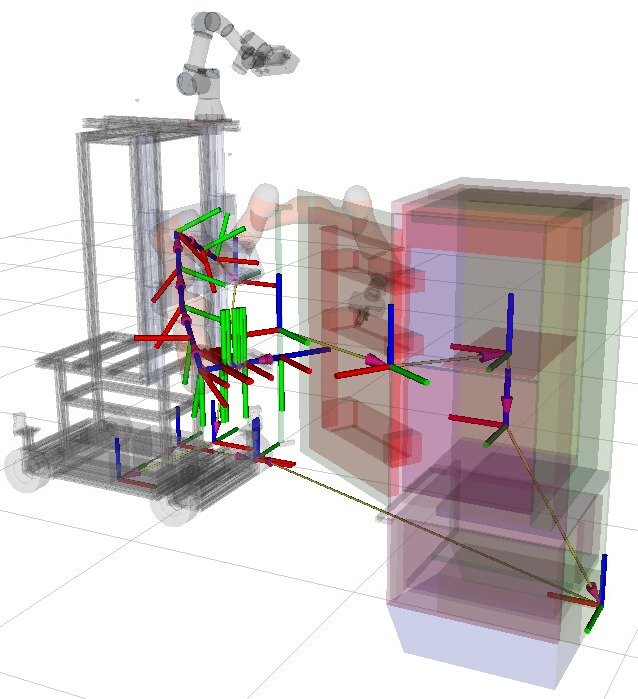}  
    &

    \includegraphics[height=33ex]%
    {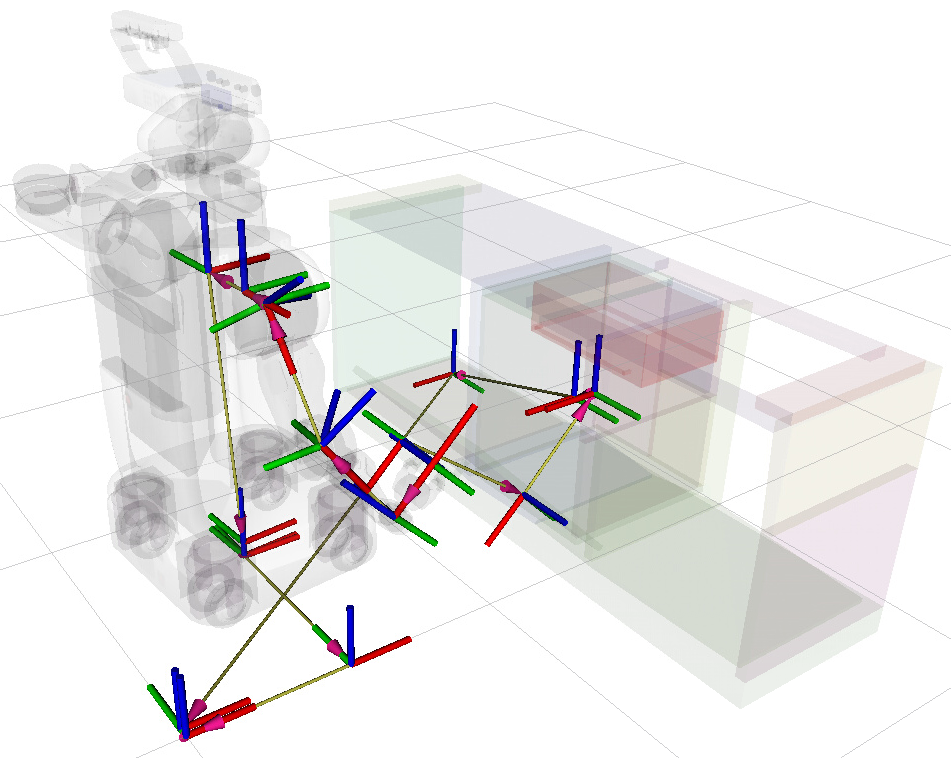}  
    &

     \includegraphics[height=33ex]%
      {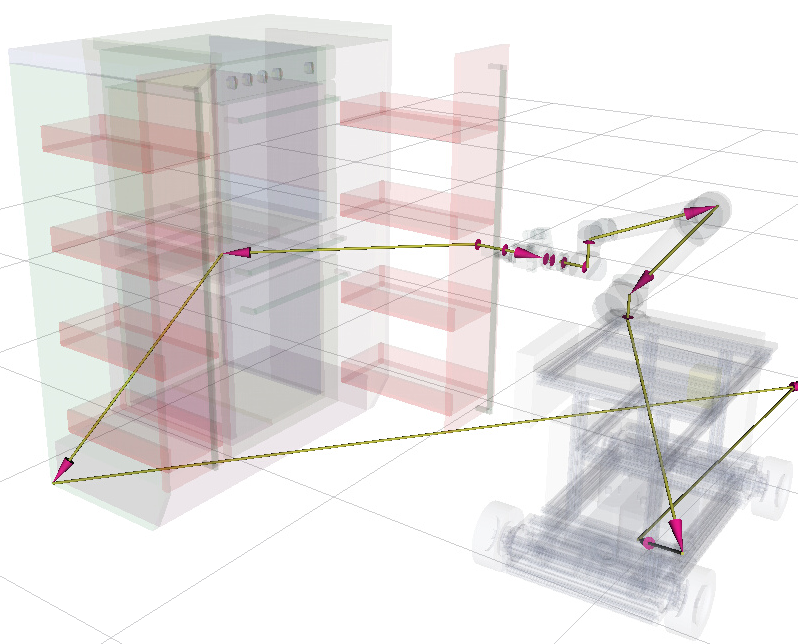}  \\
 (a) & (b) & (c)
  \end{tabular}

  \caption[Generating context-specific motions fom general motion   goals in \textsc{Giskard}]%
  {Three different robots executing the action request ``open the container holding the object to be placed on the table''  for different objects located in different containers.}
  \label{fig:opening-containers}
\vspace{-1 mm}
\end{figure*}

\subsection{Giskard: The Action Executive}
\label{sec:actionexecutive}

Given a motion plan created by the CRAM Plan Executive, the Giskard  Action Executive  provides semantic constraint- and
optimization-based motion control for manipulation actions.  Giskard can be tasked with motion goals and objectives, such as ``keep holding a door handle and move the handle according to the articulation model that the handle is part of''. These two motion objectives are sufficient to open and close all containers of the kitchen furniture: the oven, the dishwasher, the refrigerator, the drawers, and the cupboards. Giskard can also execute these motion objectives for different robots; see Fig. \ref{fig:opening-containers}.

As constraint- and optimization-based control is a mathematical optimization problem, Giskard transforms object-based action and motion specifications into mathematically formalized motion tasks. 
While the optimization-based motion generation can compute good body motions, it does so based on idealized and abstract models of the robot capabilities. In many situations the model assumptions are not satisfied. This happens, for example, if the robot moves different body parts at the same time and the composed motion causes inaccuracies in hand motions that are too large for grasping objects successfully, or the motion generation does not take into account the inaccuracies of the estimation of the robot pose. Such motion generation problems can often be better approached through experience-based learning, for example, by learning manipulation strategies using reinforcement learning. This has been investigated for learning hand manipulation strategies to open and close containers using tactile-based manipulation procedures which are linked to the declarative aspects of the developed tactile state detections \cite{meier2020from}. Such processes include movement guidance, tactile servoing, tactile exploration with respect to shape or moveability, and different forms of task-related force and touch-based control, e.g.  when unscrewing a lid or cutting a piece of bread.  At the more abstract semantic level, the procedures need to include concepts which cover processes that extend over a range of several seconds, while incorporating their declarative and procedural aspects in a low dimensional PEAM representation, a concept described in the next section.

\subsection{COGITO: The Metacognition System}
\label{sec:metacognition}
  
Metacognition is an ability to reason about the performance of the system and adapt the cognitive architecture to improve its performance.  In the CRAM cognitive architecture, this translates to an ability to reason about the generalized action plans, the plan interpreter, and the generative model encapsulated in the KnowRob 2.0  KR\&R system, and then self-program to adapt its capabilities.  This self-programming precondition of metacognition in  the CRAM cognitive architecture is satisfied because CPL is an extension of the Lisp programming language.  There are two properties of the Lisp language that facilitate metacognitive capabilities: (i) programs as data,  and (ii) the existence of metacircular interpreters. In Lisp, programs are represented as nested lists, that is as Lisp data structures. This means that Lisp programs can inspect and modify themselves. In other words, CPL plans can inspect and modify CPL plans. The idea of metacircular interpretation is that an interpreter for a programming language can be implemented in the language itself. This metacircular interpretation process can be used to make the interpretation of a robot control program explicit and represent it for introspection and meta-reasoning.  In other words, the CPL interpreter can inspect and modify itself.

The plan executive and the action plans are designed  to facilitate introspective reasoning: the inference and perception tasks are
represented in the plan in a modular and transparent form and  the queries generated during plan execution and their answers are recorded together with the success and failure of the corresponding action. This provides the robot with a form of computational awareness: the means to reason about the inferences (and the reasoning that produces them) when inferring  what constitutes appropriate context-specific behavior. Consequently, the robot can answer queries about the inference tasks it needs to solve to perform an action, about any information that is missing when determining the appropriate behavior, about the inferred proposed behavior,  about whether  the  behavior would achieve the desired outcome, and about any unwanted side effects. In the CRAM cognitive architecture, this awareness of body motion reasoning is an essential factor for the cognitive development of everyday manipulation capability because it enables the robot agent to assess the reasoning mechanisms  and substitute inference mechanisms with better ones, if necessary. Thus, the robot can improve its manipulation capability by improving its reasoning capability. 

Introspective reasoning enables robot agents to answer questions regarding to \emph{why} the robot made certain decisions, \emph{why} it holds certain beliefs, and \emph{why} it believes that certain physical events occurred.  We adopt Feynman's view that answering ``why'' only makes sense relative to a body of knowledge that is accepted as being true. Without such an asserted truth, asking repeatedly ``why'' is a process that does not terminate and is sometimes even circular. In the CRAM cognitive architecture,  this accepted body of knowledge is the \textsc{SOMA} ontology: the models of the environment, the body of the robot, the data structures and computational processes, and  the control program itself, along with the body motions and their physical effects are all formalized in the ontology. The fact that these assertions are logical axioms lets the robot agent automatically infer all the implicit knowledge implied by the ontology and use this knowledge for introspection.

Meta-reasoning leverages an important aspect in recording NEEMs. Specificially, it leverages the fact that the robot agent segments motions into submotions,  decomposes the interpretation of plans into the interpretations of subplans, and  asserts relations
between them. For example, the robot agent might assert that a motion phase of the motion plan has generated an episode of body motion, and that this episode of body motion has caused a change in the environment. This representational structure thus allows the robot agent to identify the subplan that is responsible for the outcome of an action, e.g. opening a drawer. The ability to make such inferences --- to map from things that have happened to the process that caused them to happen --- is the key to making targeted changes to the plan and  allowing sophisticated self-programming to improve the robot agent's cognitive abilities. For example, consider the situation where, in interpreting a generalized action plan through the process of contextualization, sampling the generative model does not yield valid motion plan parameter values. Giving the robot agent the ability to reprogram the generalized action plan for specific task variations and contexts provides the means to overcome this impasse by trying new actions, e.g. closing a door by pushing with the elbow or the foot instead of grasping the handle and moving the hand according to the articulation model of the door. The generative model facilitates this reprogramming through transformational planning and learning and, as described in Section \ref{section:generative-model}, we have  implemented the first limited realization of such transformational learning: revising generalized plans to change the activity structure for specific task and context variations \cite{kazhoyan20transformation}.  

We return to the discussion of metacognition in Section \ref{section:discussion} where we outline planned extensions to the CRAM cognitive architecture.

\section{Demonstration of the CRAM-controlled Robot Carrying Out Everyday Activities}
\label{section:demo}

The current abilities of CRAM  to accomplish everyday activities has been demonstrated in an extensive validation exercise requiring a robot agent to set  a table for a meal and and clear up afterwards, given underdetermined task requests. The operation of the robot when carrying out these activities is captured on video (see Footnote \ref{footnote:video}). The approach we have adopted exploits the \textsf{\footnotesize fetch\&place} generalized action plan. As we have noted, this plan can be executed with different types of robots and it can be used with various objects and in different environments. The plan is automatically  contextualized for each individual object transportation task. Thus, the robot  infers the body motions required to achieve the respective object transportation task depending on the type and state of the object to be transported (be it a spoon, bowl, cereal box, milk box, or mug), the original location (be it the drawer, the high drawer, or the table), and the task context (be it setting or cleaning the table, loading the dishwasher, or throwing away items) \cite{kazhoyan21easemilestone}. In doing this, it avoids unwanted side effects (e.g. knocking over a glass when placing a spoon on the table). The body motions generated to perform the actions are varied and complex and, when required, include subactions such as opening and closing containers, as well as coordinated, bimanual manipulation tasks.  Fig. \ref{fig:blackbox-mastery}  shows some examples of the robot grasping objects in different ways during this exercise. All of these grasp strategies are inferred from the robot's knowledge base: the spoon in the drawer is grasped from the top because it is a very flat object; the tray is grasped with two hands because the center of mass would be too far outside the hand for a single-hand grasp; the mug is grasped from the side and not at the rim because the purpose of grasping it is pouring liquid into the mug.

\section{Discussion and Future Extensions}
\label{section:discussion}

The long term vision is for CRAM to exploit transformational learning and self-programming in two complementary ways: by specialization   and by generalization. Both approaches are discussed in the following two paragraphs. While they  represent future extensions to the CRAM cognitive architecture,  they are solidly based on the computational foundations described above.

Beneath the familiarity of everyday activities often lies a complexity that can be computationally intractable, especially when you factor in the flexibility that humans exhibit when carrying out these activities, the variety of circumstances in which they carry them out, and the underdetermined manner in which they are described.  This complexity is characterized by the fact that the mapping encapsulated in the CRAM generative model is embedded in a very high dimensional space, i.e.  mapping from a vaguely-stated high-level goal to the specific low-level motion parameter values required to accomplish the action successfully.
One of the central ideas of the CRAM cognitive architecture is that, for everyday activities, the generative model does not need to capture all the dimensions of this space:  subsets of these dimensions are often sufficient to accomplish the actions successfully. These subsets are manifolds, specifically PEAMs (pragmatic everyday activity manifolds), and they serve to render tractable the  solution of problems that in their full generality are intractable. A PEAM, therefore, represents a form of specialization.  It is achieved using the constraints that knowledge of everyday activities and the environment bring to bear on the problem.\footnote{The idea of exploiting subspace manifolds to render an otherwise intractable problem tractable has parallels in other related domains. For example, in the context of dynamical systems, \cite{Schoener09} argues that it is possible for a dynamical system model to capture the behavior of a very high dimensional connectionist system using a small number of variables because the macroscopic states of high-dimensional dynamics and their long-term evolution are captured by the dynamics in that part of the space where instabilities occur, known as the low-dimensional center-manifold.} One of the main goals of CRAM is to identify and exploit the PEAMs that will result in a robot agent mastering everyday activities.

Generalization through metacognitive induction complements the PEAM solution strategy by exploring patterns among generalized actions plans, seeking ways to transform generalized action plans, either by carrying out the action in a more efficient and effective manner, or by accomplishing the outcome of the action in a different way. For example, instead of depending on a generalized action plan to carry dishes one by one to the dishwasher, a more general plan might first stack them if, as in the case of plates, they are stackable, then carry them together,  and transfer them from the stack into the dishwasher. Alternatively, if they are not stackable, they might be placed on a tray, carried, and transferred to the dishwasher from the tray.  The embedding of  this form of induction and transformational learning in the CRAM metacognition system is also one of the main goals for the future.

\begin{figure*}[tb]
  \begin{center}
    \includegraphics[width=0.8\textwidth]%
    {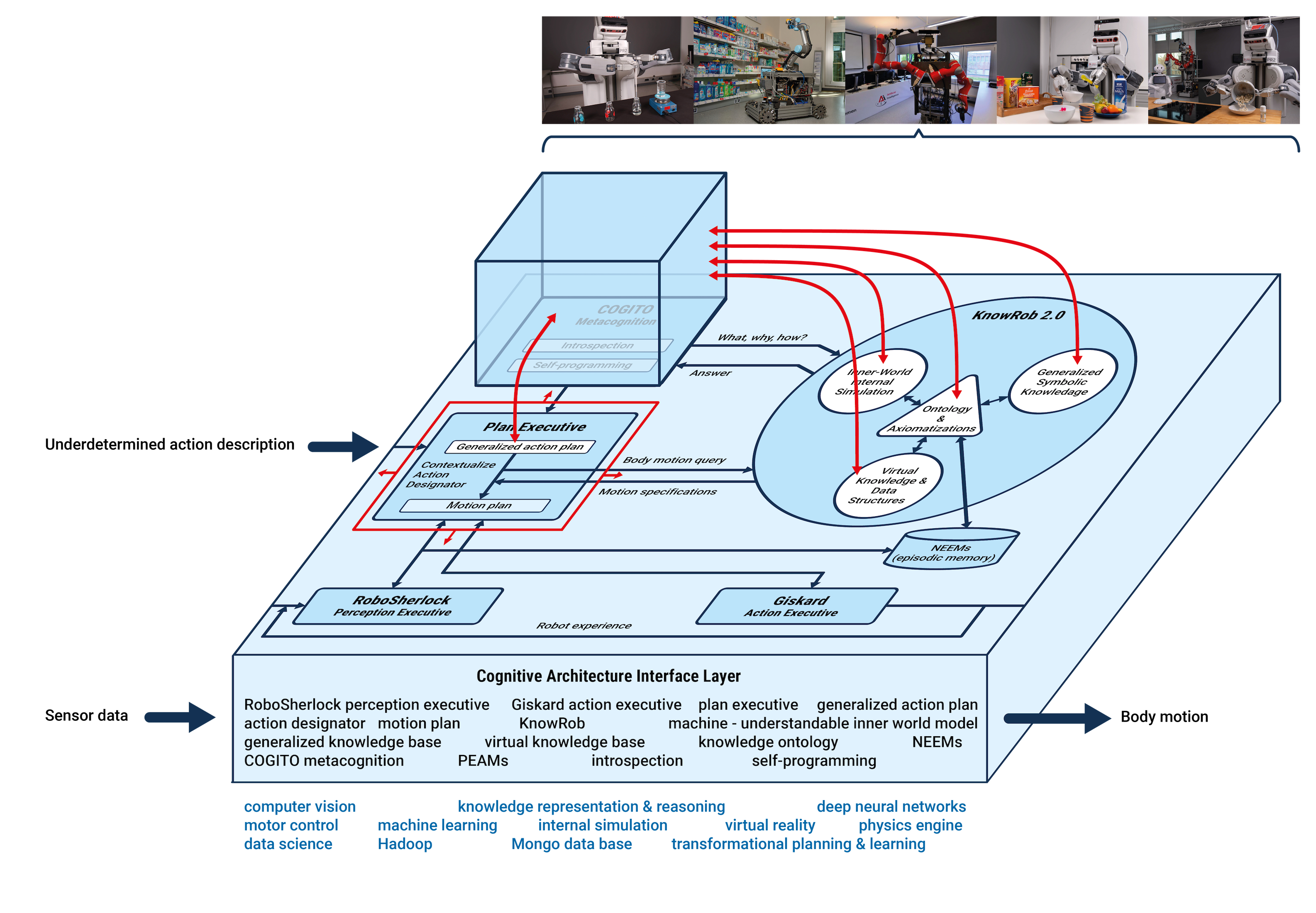}
    \caption[The CRAM cognitive architecture as an interface layer]{The cognitive architecture viewed as an interface layer exposing a self-programmable generalized action plan interpreter that can be reprogrammed by the metacognition system (above the interface) to effect generalization and specialization of plans. This results in an extended generalized plan language (exposed by  the interface and depicted in white) and an extended interpreter (shown by the red rectangle).  Abstract knowledge in the knowledge ontology, the machine-understandable inner-world model, the virtual knowledge base, and the generalized knowledge base (all exposed by the interface and depicted in white) are updated accordingly.  This abstract interface allows different robots to be used for different tasks in different domains.  All the implementation processes and representations associated with the contextualization of the generalized plan language are hidden below the interface layer.}
    \label{fig:interface-layer}
  \end{center}
\vspace{- 5mm}
\end{figure*}

We can re-frame this metacognition --- specialization and generalization --- in the following general manner. 

The performance of  underdetermined action descriptions or, equivalently, the interpretation of generalized action plans, requires a programming language and interpreter to generate robot body motions for  the associated manipulation tasks.  This programming language must leverage abstract knowledge, allowing the robot to understand what it is doing, accomplish novel manipulation tasks, and learn from very few examples. Our stance is that  the cognitive architecture is an extensible interpreter for this language and that it can be viewed as the interface layer beneath which lie the complex mechanisms that (a) map instantiated generalized action plans to likely-to-succeed motion parameter values and (b) execute these motions adaptively; see Fig. \ref{fig:interface-layer}. Above this interface layer lie the mechanisms by which generalized action plans are flexibly instantiated and extended. 

The cognitive mechanisms operate on the joint distribution of motions and their effects, of which there will be  many (at least one for each action category). These joint distributions are generative internal models which, when subjected to the constraints imposed by reasoning with contextual knowledge, yield the sample in the joint distribution  that maximizes the likelihood that the action will succeed.  This sample has the maximum expected success measure over all internal models that are relevant in the current context.    These generative internal models can be learned from experience and are composable so that they can be recombined  to yield novel action strategies.  

The interface layer, i.e.\ the cognitive architecture,  interprets the generalized action plan by executing the three steps of contextualization and then, by deploying the action executive, adaptively executing the parameterized motion plan using 
parameter values produced in the third step in the contextualization process. However, because the cognitive architecture is a plan interpreter, it is also a  symbolic program. This means that  metacognitive processes above the interface layer can re-program the intrepreter to extend the general action plan language it interprets. This extension improves the ability of the cognitive 
architecture to identify the robot body motions that are likely to succeed in accomplishing an underdetermined action description.  
This extended language expresses new generalized  action plans which are generated by the metacognitive processes that implement 
transformational planning and learning.  The metacognitive processes also generate  specialized action plans that exploit PEAMs  to achieve feasible solutions to otherwise intractable problems by identifying the constraints that knowledge of everyday activies and the environment can bring to bear; see Section \ref{sec:metacognition}.

The CRAM cognitive architecture makes it possible to implement  metacognition in a unique way. While most cognitive architectures view metacognition as a separate independent module responsible for the oversight of the performance of the cognitive architecture, the CRAM  cognitive architecture can implement the metacognitive functions (of plan generalization, plan specialization, and extension of the plan language and plan language interpreter) using the plan executive itself, recruiting the KR\&R executive to effect the necessary inference mechanisms.  It can achieve this by exploiting the virtual knowledge base in the KnowRob 2.0 KR\&R framework; see Figure~\ref{fig:cram-cognitive-architecture}.  This knowledge base provides  a view (in the technical sense) of the plan executive, i.e.\ a dynamically-instantiated abstract representative of the CRAM plan language interpreter. This virtual knowledge is also axiomatized in the knowledge ontology to expose the semantics of the interpreter.  This means that the knowledge  representation and reasoning executive can then be used by the plan executive to reason about itself and thereby achieve the metacognitive functions mentioned above.  In this way, the plan executive can effectively self-program and metacognition can be considered to be a logical extension of the plan executive with re-entrant processing.

Once these three planned developments have been implemented, the plan executive will then have three distinct responsibilites: (i)
plan execution via contextualization using the generative model, (ii) plan recovery via plan monitoring and failure handling, and (iii) plan language extension via metacognition, i.e. plan generalization and plan specialization.

\section{Summary and Conclusions}
\label{section:robotic_agency_summary}

The CRAM cognitive architecture is founded on the concept of a generative model and a conceptual  framework for accomplishing everyday manipulation tasks.  These are based on the following three hypotheses:
\begin{enumerate}
\item We can provide for each manipulation action category such as \textsf{\footnotesize fetch}, \textsf{\footnotesize place}, \textsf{\footnotesize pour}, \textsf{\footnotesize cut}, \textsf{\footnotesize wipe} a general motion plan schema with motion phases and phase-specific motion parameters that can generate a range of body motions to achieve the respective goal of the action in a large variety of contexts.
\item A request for performing actions can be represented by  declarative, symbolic, and underdetermined action descriptions that are to be refined through knowledge, reasoning, and perception in  order to infer the motion parameter values that generate a body motion  to achieve the goal of the action description.
\item The knowledge needed to refine action descriptions can be  represented as generalized and modular knowledge chunks that can be  composed through a reasoning engine to derive appropriate motion parameterizations for novel tasks, situations, and contexts. 
\end{enumerate}
To perform an underdetermined action description, the CRAM cognitive architecture contextualizes  it by (a) instantiating the associated generalized action plan,  (b) extending it by adding the parameters needed to execute the motion plan,  and (c) creating a query for the parameter values that maximize the likelihood that the associated body motions  will successfully accomplish the desired action. The action executive then adaptively executes the body motions, as demonstrated in Section \ref{section:demo}.

We conclude by re-asserting the power of viewing the implicit-to-explicit execution of an underdetermined action description as a sampling of a joint probability distribution of (a) motions that the generalized action plan generates and (b) the physical effects these motions cause, such that the sample values maximize the likelihood of success in carrying out the action.  The CRAM approach leverages explicitly-represented knowledge  \& reasoning and inner-world simulation-based prospection in order to sample this joint probability distribution. It does not use an explicit graphical model or Bayesian network to infer the motion parameter values but it does not preclude it either. However, the joint probability distribution  allows us to conceptualize what we seek in the knowledge-based contextualization process: to maximize the utility of the selected motion parameter values, i.e. to maximize the likelihood of the action being successful. This corresponds to a probability distribution that has low entropy and is highly informative: one that exhibits sharp peaks across the probability distribution, indicating motion parameter values that have a high probability  of achieving the desired outcome.  

It is the exploitation of knowledge that shapes the probability distribution in this manner and decreases the entropy of the distribution, tuning the distribution to the context based on  knowledge of the environment in which the activity is being conducted, knowledge provided  both by perception and by prospection through inner-world high-fidelity virtual reality and physics engine simulation.   Thus, when querying the KnowRob~2.0\ knowledge base for the motion parameter values that are most likely to succeed in accomplishing the desired action outcome,  reasoning, perception, and prospection together yield a probability distribution  with sharp peaks, and  the parameter values corresponding to the peak with the highest probability are selected as the response to the query.  By analogy with Bayesian reasoning, CRAM uses evidence supplied by prior knowledge, reasoning, perception, and prospection  to select the motion plan parameter values that are maximally-likely to result in a successful action.
 
Ultimately, the CRAM cognitive architecture can be viewed as an interface layer that, by virtue of its extensible Lisp-based CRAM plan language interpreter, is well-suited to autonomous development through self-programming. At present, the CRAM cognitive architecture does not yet have this developmental capacity but it has demonstrated the practical realization of cognition-enabled robotics and its application in carrying out everyday activities, and it provides both the principled foundations and the practical computational means to pursue this challenging research goal to a successful conclusion, endowing robots with a capacity for flexible, context-sensitive manipulation in everyday activities.

\section*{Acknowledgment}
The authors would like to thank the many people who have contributed to the development of the CRAM cognitive architecture over the past several years.  These include: 
Ferenc B\'{a}lint-Bencz\'{e}di,
Georg Bartels,  
Daniel Be{\ss}ler,
Andrei Haidu, 
Franklin Kenghagho Kenfack,
Sebastian Koralewski,
Lorenz M\"osenlechner,
Daniel Nyga,
Simon Stelter,
and
Moritz Tenorth.

The preparation of this article has been  supported by the German Research Foundation DFG, as part of Collaborative Research Center (Sonderforschungsbereich) 1320 “EASE - Everyday Activity Science and Engineering”, University of Bremen (http://www.ease-crc.org/)

\ifCLASSOPTIONcaptionsoff
  \newpage
\fi



\bibliographystyle{IEEEtran}
\bibliography{cognitive_systems}
%

%

\vspace{-10 mm}
\begin{IEEEbiography}[{\includegraphics[width=1in,height=1.25in,clip,keepaspectratio]{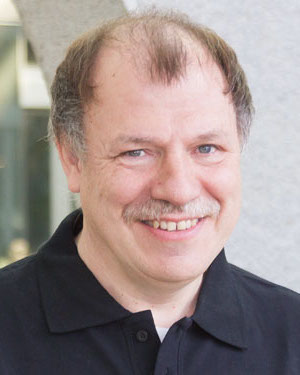}}]{Michael Beetz}  is a professor for Computer Science at the Faculty for Mathematics \& Informatics, University Bremen, and head of the Institute for Artificial Intelligence. He received his diploma degree in Computer Science with distinction from the University of Kaiserslautern. His MSc, MPhil, and PhD degrees were awarded by Yale University in 1993, 1994, and 1996, and his Venia Legendi from the University of Bonn in 2000. Michael Beetz was vice-coordinator of the German cluster of excellence CoTeSys (Cognition for Technical Systems). He is the coordinator of the German collaborative research centre EASE (Everyday Activity Science and Engineering), since 2017.
\end{IEEEbiography}
\vspace{-10 mm}
\begin{IEEEbiography}[{\includegraphics[width=1in,height=1.25in,clip,keepaspectratio]{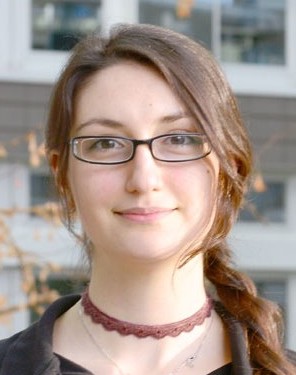}}]{Gayane Kazhoyan} is a PhD student at the Institute for Artificial Intelligence, University of Bremen. Her main research interests are concentrated in the area of cognition-enabled robot executives. She is the current lead developer of the CRAM reactive planning and execution framework. Before joining IAI in 2013, she worked as a research assistant at Kastanienbaum GmbH (now Franka Emika GmbH), in collaboration with DLR. There she was mainly working on impedance control, system identification and application programming. Before that she has acquired her M.Sc. degree in Informatics with a major in AI and Robotics at the Technical University of Munich.
\end{IEEEbiography}
\vspace{-10 mm}
\begin{IEEEbiography}[{\includegraphics[width=1in,height=1.25in,clip,keepaspectratio]{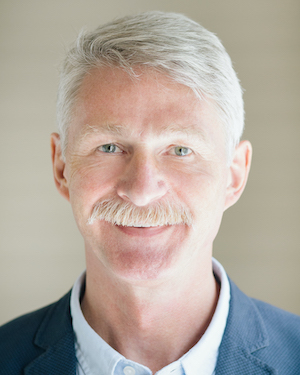}}]{David Vernon} is a researcher at the Institute for Artificial Intelligence, University of Bremen. He graduated with B.A., B.A.I. degrees in engineering from Trinity College Dublin in 1979 and joined Westinghouse Electric as a software engineer. He completed a Ph.D. at Trinity College Dublin in 1985.  Before joining the University of Bremen, he was as a professor at Carnegie Mellon University Africa, Rwanda.  He is a Fellow of the Institution of Engineers of Ireland, Senior Member of the IEEE, Research Fellow at the Kigali Collaborative Research Centre, Rwanda. He is co-chair of the IEEE RAS Technical Committee for Cognitive Robotics.  
\end{IEEEbiography}






\end{document}